\pdfoutput=1
\documentclass[11pt]{article}

\usepackage{emnlp2021}

\usepackage{times}
\usepackage{latexsym}

\usepackage[T1]{fontenc}
\usepackage[utf8]{inputenc}

\usepackage{microtype}

\usepackage{graphicx}
\usepackage{comment}
\usepackage{amsfonts}
\usepackage{amsmath}
\usepackage{subfig}
\usepackage{tabularx} 
\usepackage{mathrsfs}
\usepackage{colortbl}
\usepackage{multirow}
\usepackage[font=small,labelfont=bf]{caption}
\usepackage{footmisc}
\usepackage{wrapfig,lipsum,booktabs}
\usepackage{mathrsfs}

\usepackage{hyperref}
\hypersetup{
	colorlinks=true,
	linkcolor=blue,
}

\newcolumntype{x}[1]{>{\centering\arraybackslash}p{#1pt}}

\newlength\savewidth\newcommand\shline{\noalign{\global\savewidth\arrayrulewidth
		\global\arrayrulewidth 1pt}\hline\noalign{\global\arrayrulewidth\savewidth}}
\newcommand{\tablestyle}[2]{\setlength{\tabcolsep}{#1}\renewcommand{\arraystretch}{#2}\centering\small}
\newcommand{\ie}{\textit{i}.\textit{e}.}
\newcommand{\eg}{\textit{e}.\textit{g}.}
\title{On Pursuit of Designing Multi-modal Transformer for Video Grounding}

\author{Meng Cao\textsuperscript{1}, Long Chen\textsuperscript{2}\thanks{\ \ Work started when Long Chen at Tencent AI Lab.}, Mike Zheng Shou\textsuperscript{3}, Can Zhang\textsuperscript{1}, and Yuexian Zou\textsuperscript{1,4}\thanks{\ \ Corresponding author.}\\
\textsuperscript{1}School of Electronic and Computer Engineering, Peking University\\ 
\textsuperscript{2}Columbia University
\textsuperscript{3}National University of Singapore
\textsuperscript{4}Peng Cheng Laboratory\\
{\tt mengcao@pku.edu.cn, zjuchenlong@gmail.com}\\
{\tt {mike.zheng.shou@gmail.com}, \tt \{zhangcan, zouyx\}@pku.edu.cn }\\
}

\begin{document}
\maketitle

\begin{abstract}\label{abs}
	Video grounding aims to localize the temporal segment corresponding to a sentence query from an untrimmed video. Almost all existing video grounding methods fall into two frameworks: 1) Top-down model: It predefines a set of segment candidates and then conducts segment classification and regression. 2) Bottom-up model: It directly predicts frame-wise probabilities of the referential segment boundaries. However, all these methods are not \emph{end-to-end}, \ie, they always rely on some time-consuming post-processing steps to refine predictions. To this end, we reformulate video grounding as a set prediction task and propose a novel end-to-end multi-modal Transformer model, dubbed as \textbf{GTR}. Specifically, GTR has two encoders for video and language encoding, and a cross-modal decoder for grounding prediction. To facilitate the end-to-end training, we use a \emph{Cubic Embedding} layer to transform the raw videos into a set of visual tokens. To better fuse these two modalities in the decoder, we design a new \emph{Multi-head Cross-Modal Attention}. The whole GTR is optimized via a \emph{Many-to-One} matching loss. Furthermore, we conduct comprehensive studies to investigate different model design choices. Extensive results on three benchmarks have validated the superiority of GTR. All three typical GTR variants achieve record-breaking performance on all datasets and metrics, with several times faster inference speed. Our project is available at \href{https://sites.google.com/view/mengcao/publication/gtr}{GTR}.
	% We hope our GTR and exploration findings can help to guide the multi-modal Transformer designs in other tasks.
\end{abstract}

\section{Introduction}\label{sec:intro}
%-----------------------------------------------------------------------
\begin{figure}[t]
	\centering
	%\begin{subfigure}[b]{0.5\textwidth}
	%\centering
	%\includegraphics[width=0.5\textwidth]{figs/teaser.pdf}
	%\text{(a)}
	%\vspace{-5mm}
	%\caption{(a)}
	%\end{subfigure}
	%\begin{subfigure}[b]{0.5\textwidth}
	\centering
	\includegraphics[width=0.45\textwidth]{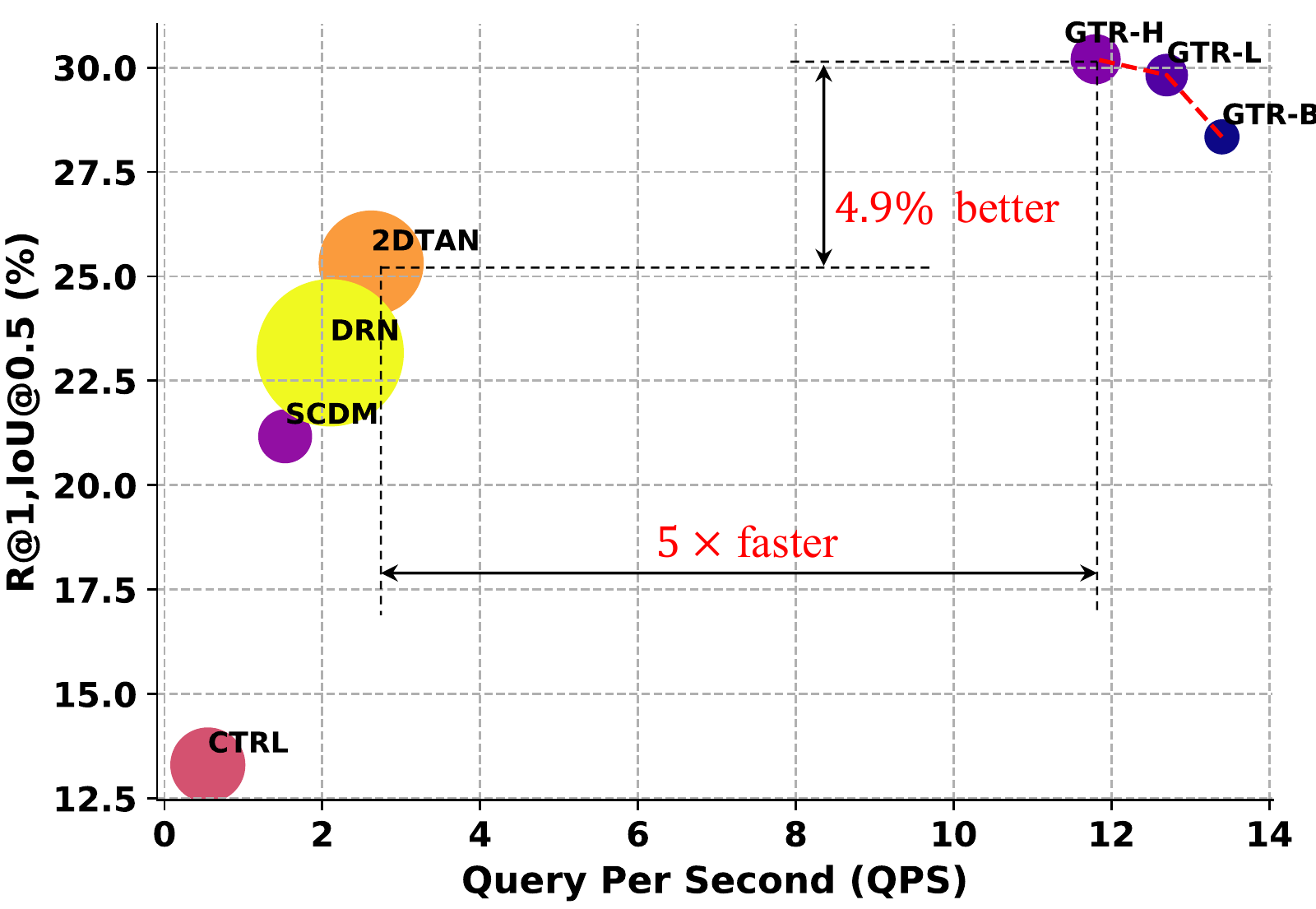}
	%\text{(b)}
	\vspace{-3mm}
	\caption{\footnotesize{Performance comparisons on TACoS in terms of R@1, IoU@0.5 and Query Per Second (the number of queries that are retrieved each second during inference). Marker sizes are proportional to the model size. Our GTR-H is 4.9\% better than 2D-TAN~\cite{zhang2020learning} with 5 times faster speed.}}
	\label{fig:teaser}
	\vspace{-3mm}
\end{figure}
%-----------------------------------------------------------------------
Video grounding is a fundamental while challenging task for video understanding and has recently attracted unprecedented research attention~\cite{chen2018temporally,chen2019weakly,chen2019localizing,zhang2019man,liu2018attentive,yuan2021closer}. Formally, it aims to identify the two temporal boundaries of the moment of interest based on an input untrimmed video and a natural language query. Compared to the conventional video action localization task~\cite{shou2016temporal,zhao2017temporal,zhang2021cola},  video grounding is more general and taxonomy-free, \ie, it is not limited by the predefined action categories.

The overwhelming majority of the state-of-the-art video grounding methods fall into two frameworks: 1) Top-down models~\cite{anne2017localizing,gao2017tall,ge2019mac,chen2018temporally,zhang2019cross,liu2018cross,yuan2019semantic}: They always use a propose-and-rank pipeline, where they first generate a set of moment proposals and then select the best matching one. To avoid the proposal bottlenecks and achieve high recall, a vast number of proposals are needed. Correspondingly, some time-consuming post-processing steps (\eg, non-maximum suppression, NMS) are introduced to eliminate redundancy, which makes the matching process inefficient (cf. Figure~\ref{fig:teaser}). 2) Bottom-up models~\cite{mun2020local,rodriguez2020proposal,zeng2020dense,chen2020rethinking,lu2019debug}: They directly regress the two temporal boundaries of the referential segment from each frame or predict boundary probabilities frame-wisely. Similarly, they need post-processing steps to group or aggregate all frame-wise predictions.

Although these two types of methods have realized impressive progress in video grounding, it is worth noting that they still suffer several notorious limitations: 1) For top-down methods, the heuristic proposal generation process introduces a series of hyper-parameters. Meanwhile, the whole inference stage is computation-intensive for densely placed candidates. 2) For bottom-up methods, the frame-wise prediction manner overlooks fruitful temporal context relationships, which strictly limits their performance. 3) All these methods are not \emph{end-to-end}, which need complex post-processing steps to refine predictions, and easily fall into the local optimum.

%To simplify the pipeline and bypass these surrogate tasks, we aim at removing unnecessary designs by making the training objective and output results direct and transparent. Recently, DETR~\cite{carion2020end}, a Transformer-based detector has explored to address this issue in the field of object detection. However, extending DETR to video grounding is non-trival since it requires to capture video temporal correlations and perform multi-modal interactive reasoning.

In this paper, we reformulate the video grounding as a set prediction problem and propose a novel end-to-end multi-modal Transformer model \textbf{GTR} (video \textbf{G}rounding with \textbf{TR}ansformer). GTR has two different encoders for video and language feature encoding, and a cross-modal decoder for final grounding result prediction. Specifically, we use a \emph{Cubic Embedding} layer to transform the raw video data into a set of visual tokens, and regard all word embeddings of the language query as textual tokens. Both these visual and textual tokens are then fed into two individual Transformer encoders for respective single-modal context modeling. Afterwards, these contextualized visual and textual tokens serve as an input to the cross-modal decoder. Other input for the decoder is a set of learnable segment queries, and each query try to regress a video moment by interacting with these contextualized tokens. To better fuse these two modalities in the decoder, we design a new \emph{Multi-head Cross-Modal Attention} module (MCMA). The whole GTR model is trained end-to-end by optimizing a \emph{Many-to-One Matching Loss} which produces an optimal bipartite matching between predictions and ground-truth. Thanks to this simple pipeline and the effective relationship modeling capabilities in Transformer, our GTR is both effective and computationally efficient with extremely fast inference speed (cf. Figure~\ref{fig:teaser}).

Since our community has few empirical experiences on determining the best design choice for multi-modal Transformer-family models, we conduct extensive exploratory studies on GTR to investigate the influence of different model designs and training strategies, including: (a) \emph{Visual tokens acquisition}. We use a cubic embedding layer to transform raw videos to visual tokens, and discuss the design of the cubic embedding layer from three dimensions. (b) \emph{Multi-modal fusion mechanism}. We propose six types of multi-modal fusion mechanisms, and compare their performance and computation cost thoroughly. (c) \emph{Decoder design principles}. We explore some key design principles of a stronger multi-modal Transformer decoder, such as the tradeoff between depth and width, or the attention head number in different layers. (d) \emph{Training recipes}. We discuss the influence of several training tricks. We hope our exploration results and summarized take-away guidelines can help to open the door for designing more effective and efficient Transformer models in multi-modal tasks.

% (a) Visual Token Acquisition. Efficient processing of video data is the first thing worth discussing. In this respect, we propose a cubic embedding layer tailored for our transformer architecture and explore several key designs to boost its performance. (b) Multi-modal fusion mechanisms. MCMA is designed for modality fusion in the decoder and we provide several specific instantiations of it. Detailed comparisons regarding performance and computation overhead are presented. (c) Decoder Design Principles. We explore some empirical rules of the decoder design to make it effective in our video grounding task. (d) Training recipe. Conceptually, Vision Transformer requires large-scale datasets for training. To settle this issue, we discuss several notable training tricks which are important in training an effective Transformer. 

In summary, we make three contributions in this paper:
\vspace{-0.6em}
\begin{enumerate}
	\itemsep-0.4em
	\item We propose the first end-to-end model GTR for video grounding, which is inherently efficient with extremely fast inference speed.
	
	\item By the careful design of each component, all variants of GTR achieve new state-of-the-art performance on three datasets and all metrics.
	
	\item Most importantly, our comprehensive explorations and empirical results can help to guide the design of more multi-modal Transformer-family models in other multi-modal tasks.
\end{enumerate}

\section{Related Work}\label{relatedwork}
%-----------------------------------------------------------------------
\begin{figure*}[t]
	\centering
	\includegraphics[width=1.0\textwidth]{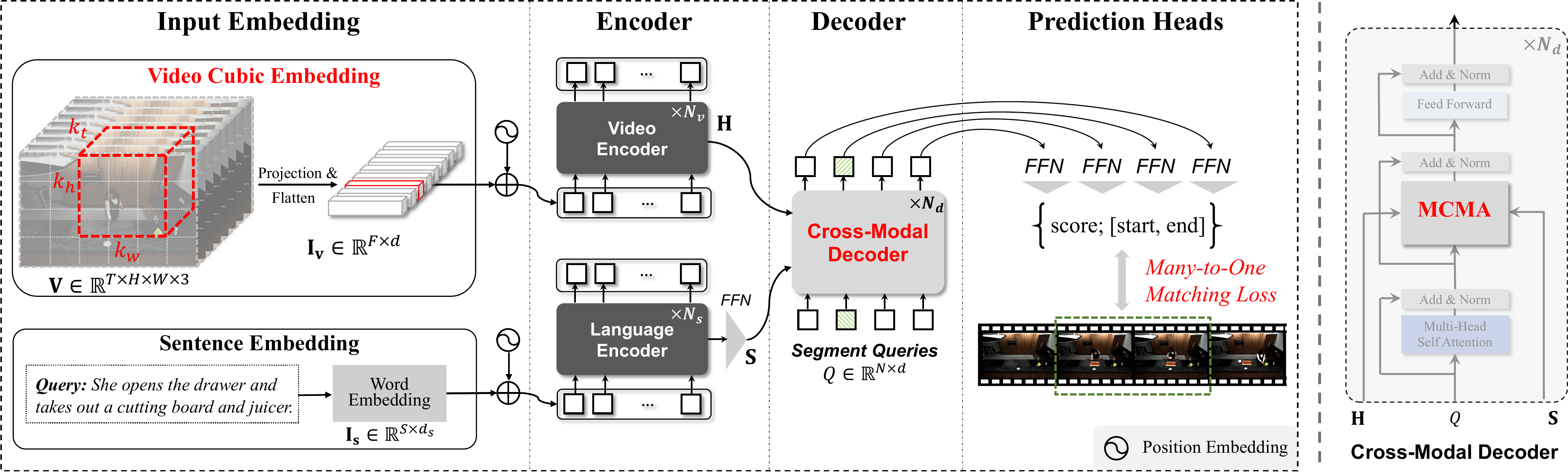}
	\vspace{-6mm}
	\caption{\footnotesize{An overview of GTR. (1) Input Embedding transforms the video and language data to feature space. (2) Encoder is applied to encode global context. (3) Decoder contains a novel Multi-head Cross-Modal Attention module (MCMA) to fuse two modalities. (4) Prediction Heads generate grounding results optimized by a Many-to-One Matching Loss.}}
	\vspace{-2mm}
	\label{fig:pipeline}
\end{figure*}

\noindent \textbf{Video Grounding.} The overwhelming majority of state-of-the-art video grounding methods are top-down models~\cite{anne2017localizing,gao2017tall,ge2019mac,liu2018attentive,liu2020jointly,zhang2019man,zhang2020learning,chen2018temporally,yuan2019semantic,wang2020temporally,xu2019multilevel,xiao2021boundary,xiao2021natural,liu2021context}. Although these top-down models have dominated the performance, they suffer from two inherent limitations: 1) The densely placed proposal candidates lead to heavy computation cost. 2) Their performance are sensitive to the heuristic rules (\eg, the number and the size of anchors). Another type of methods is bottom-up models~\cite{yuan2019find,lu2019debug,zeng2020dense,chen2020rethinking,chen2018temporally,zhang2020span}. Some works~\cite{he2019read,wang2019language} resort to Reinforcement Learning to guide the boundary prediction adjustment. However, all existing methods (both top-down and bottom-up) are not end-to-end and require complex post-processing steps. In this paper, we propose a end-to-end model GTR, which directly generates predictions with ultrafast inference speed.

%Straightforward as it seems, these methods ignore the frame-wise relationships and the performance is behind the top-down counterpart. In this paper, we propose a postprocessing-free pipeline based on a transformer architecture with ultrafast inference speed.
%we formulate video grounding as a set prediction problem and directly output the final set of predictions based on a transformer encoder-decoder architecture. This postprocessing-free pipeline is rather straightforward and leads to the ultrafast inference process.
%This simple pipeline gets rid of the parameter tuning and is postprocessing-free.

%-----------------------------------------------------------------------
\noindent \textbf{Vision Transformer.} Transformer~\cite{vaswani2017attention} is a de facto standard language modeling architecture in the NLP community. Recently, a pioneering object detection model DETR~\cite{carion2020end} starts to formulate the object detection task as a set prediction problem and use an end-to-end Transformer structure to achieve state-of-the-art performance. Due to its end-to-end nature, DETR regains the CV community attention about the Transformer, and a mass of vision Transformer models have been proposed for different vision understanding tasks, such as image classification~\cite{wang2021crossformer}, object detection~\cite{carion2020end,zhu2020deformable}, tracking~\cite{meinhardt2021trackformer,xu2021transcenter,chen2021transformer,sun2020transtrack}, person re-id~\cite{he2021transreid}, image generation~\cite{jiang2021transgan}, super resolution~\cite{yang2020learning}, and video relation detection~\cite{gao2021video}. Unlike previous methods only focusing on the vision modality, our GTR is a multi-modal Transformer model, which not only needs to consider the multi-modal fusion, but also has few empirical experience for model designs.

\section{Video \underline{G}rounding with \underline{TR}ansformer}\label{methods}
%-------------------------------------------------------------------------------------%
%\input{tables_figs/figPipeline}
\begin{figure*}[t]
	\centering
%	\begin{subfigure}[b]{\textwidth}
%		\centering
%		\includegraphics[width=\textwidth]{figs/threeFuse.pdf}
%		\caption{$m=0$}
%	\end{subfigure}
%	\begin{subfigure}[b]{0.9\textwidth}
		\centering
		\includegraphics[width=0.9\textwidth]{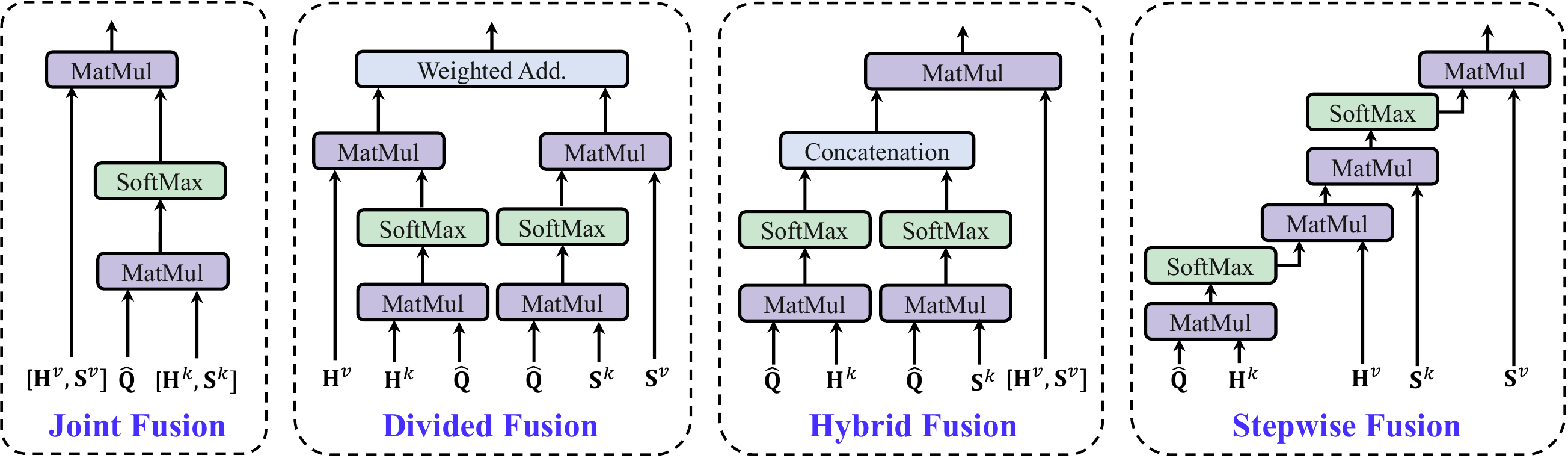}
		%\caption{$m=1$}
%	\end{subfigure}
    \vspace{-2mm}
	%\caption{\small Instantiations of  Multi-Head Cross-Modal Attention module (MCMA). There exist two variants for Hybrid Fusion and Stepwise Fusion respectively. Refer to supplementary material for detail.}
	\caption{\footnotesize{Different instantiations of the Multi-head Cross-Modal Attention module (MCMA).}}
	\vspace{-2mm}
	\label{fig:3}
\end{figure*}
%-------------------------------------------------------------------------------------%
%-------------------------------------------------------------------------------------%
%\subsection{Overview of GTR}\label{sec:3.1}
%-------------------------------------------------------------------------------------%
As shown in Figure~\ref{fig:pipeline}, GTR yields temporal segment predictions semantically corresponding to the given query by four consecutive steps: (1) \emph{Input Embedding}. Given a raw video and a query, this step aims to encode them into the feature space (\ie, visual and textual tokens). 2) \emph{Encoder}. Visual and textual token embeddings are enhanced with a standard Transformer encoder by modeling the intra-modality correlations. (3) \emph{Cross-Modal Decoder}. Contextualized visual and textual token embeddings are fused by a Multi-head Cross-Modal Attention module (MCMA), and a set of learnable segment queries are fed into the decoder to interact with these two modal features. (4) \emph{Prediction Heads}. A simple feed-forward network (FFN) is applied to predict final temporal segments.
%-------------------------------------------------------------------------------------%
\subsection{Input Embedding and Encoder}\label{sec:3.1}
\noindent \textbf{Video Cubic Embedding.} Aiming to build a pure Transformer model without the reliance on CNNs, ViT~\cite{dosovitskiy2020image} decomposes input images into a set of non-overlapping patches. To process video data, a straightforward solution is to apply this partition for each frame. However, this simple extension overlooks the frame-wise temporal correlations. Thus, we propose a Cubic Embedding layer which directly extracts 3D visual tokens from the height, width, and temporal dimensions respectively (cf. Figure~\ref{fig:pipeline}).

Formally, given a raw video, we firstly use framerate $1/\gamma_{\tau}$ to sample the video, and obtain video clip $\mathbf{V} \in \mathbb{R}^{T \times H \times W \times 3}$. Then, we use a sampling kernel $\kappa$ with shape $(k_h, k_w, k_t)$ to transform the video into visual tokens, and the sampling kernel $\kappa$ is propagated with stride size $s(s_h, s_w, s_t)$, where $s_h$, $s_w$, and $s_t$ denote the stride size in the height, width, and temporal dimension respectively. Each sampled 3D visual patch is fed into a projection layer, and the output of the cubic embedding layer is a set of visual token embeddings $\mathbf{I}_{\mathbf{v}} \in \mathbb{R}^{F \times d}$, where $F$ is the number of visual tokens, and $d$ is the dimension of the projected layer. In our experiments, we set $d$ to the same hidden dimension of the Transformer. Apparently, $F = O_h \times O_w \times O_t$ where $O_h=\left\lfloor\frac{H-k_h}{s_h}+1\right\rfloor$, $O_w=\left\lfloor\frac{W-k_w}{s_w}+1\right\rfloor$, and $O_t=\left\lfloor\frac{T-k_t}{s_t}+1\right\rfloor$. Compared to the non-overlapping tokenization manners, our cubic embedding layer allows overlapping sampling, which implicitly fuses adjacent spatial-temporal context (More experiments and discussions about the cubic embedding layer are shown in Sec.~\ref{sec:4_2}). In particular, when setting $s_h = k_h$, $s_w = k_w$, and $s_t = k_t = 1$, our cubic embedding degrades into the prevalent patch embedding in ViT.

%(ViT writings)To handle 2D images, we reshape the image x 2 RHWC into a sequence of flattened 2D patches xp 2 RN(P2C), where (H;W) is the resolution of the original image, C is the number of channels, (P; P) is the resolution of each image patch, and N = HW=P2 is the resulting number of patches, which also serves as the effective input sequence length for the Transformer. The Transformer uses constant latent vector size D through all of its layers, so weflatten the patches and map to D dimensions with a trainable linear projection (Eq. 1). We refer to the output of this projection as the patch embeddings.

\noindent \textbf{Sentence Embedding.} For the language input, we first encode each word with pretrained GloVe embeddings~\cite{pennington2014glove}, and then employ a Bi-directional GRU to integrate the sentence-level embedding feature $\mathbf{I_s} \in \mathbb{R}^{S \times d_s}$, where $S$ represents the length of the sentence query, and $d_s$ is the dimension of textual token embeddings.

%-------------------------------------------------------------------------------------%

%-------------------------------------------------------------------------------------%
\noindent \textbf{Encoder.} We use two plain Transformer encoders to model the visual and textual intra-modality context, respectively. Specifically, for the video token embeddings $\mathbf{I_v}$, we apply an encoder with $N_v$ layers to obtain their corresponding contextualized visual tokens $\mathbf{H} \in \mathbb{R}^{F \times d}$. Similarly, for the textual token embeddings $\mathbf{I_s}$, we use an encoder with $N_s$ layers to obtain contextualized textual tokens $\mathbf{S}$. Another feed-forward network with two fully-connected layers is applied to adjust $\mathbf{S}$ to be the same channel with $\mathbf{H}$, \ie, $\mathbf{S} \in \mathbb{R}^{S \times d}$.

%-------------------------------------------------------------------------------------%
\subsection{Cross-modal Decoder}\label{sec:3.2}
%-------------------------------------------------------------------------------------%
As shown in Figure~\ref{fig:pipeline}, the inputs for cross-modal decoder consist of the visual features $\mathbf{H} \in \mathbb{R}^{F \times d}$, language features $\mathbf{S} \in \mathbb{R}^{S \times d}$, and a set of learnable \emph{segment queries} $\mathbf{Q} \in \mathbb{R}^{N \times d}$. Each segment query $\mathbf{q}_i \in \mathbf{Q}$ tries to learn a possible moment by interacting with $\mathbf{H}$ and $\mathbf{S}$, and the whole decoder will decode $N$ moment predictions in parallel. For accurate segment localization, video grounding requires modeling fine-grained cross-modal relations. To this end, we design a cross-modal decoder with a novel Multi-head Cross-Modal Attention module (MCMA). As shown in Figure~\ref{fig:3}, we propose several specific instantiations of MCMA\footnote{More details are left in the supplementary materials. \label{supple}}:

%and detailed performance analysis is presented in Sec.~\ref{sec:4_2}.
%-------------------------------------------------------------------------------------
%\input{tables_figs/figModalFuse}
%-------------------------------------------------------------------------------------

\noindent \textbf{Joint Fusion.} Given visual and language features $\mathbf{H}$ and $\mathbf{S}$, we firstly generate a set of modal-specific key ($\mathbf{H}^k$, $\mathbf{S}^k$) and value ($\mathbf{H}^v$, $\mathbf{S}^v$) pairs by linear transformations: $\mathbf{H}^k=\mathbf{H} \mathbf{W}_h^k$, $\mathbf{S}^k=\mathbf{S} \mathbf{W}_s^k$, $\mathbf{H}^v=\mathbf{H} \mathbf{W}_h^v$, $\mathbf{S}^v=\mathbf{S} \mathbf{W}_s^v$, where $\mathbf{W}_h^k$, $\mathbf{W}_s^k$, $\mathbf{W}_h^v$ and $\mathbf{W}_s^v \in \mathbb{R}^{d \times d}$ are all learnable parameters. Then joint fusion concatenates the two modalities before conducting the attention computing:
\begin{equation*}
 \mathbf{Y}_{\text{joint}} = \text{MHA}(\mathbf{\hat{Q}}, \mathbf{H}^k \otimes \mathbf{S}^k, \mathbf{H}^v \otimes \mathbf{S}^v),
\end{equation*}
where $\mathbf{\hat{Q}}$ is the enhanced segment query embeddings after the self-attention and $\otimes$ denotes the channel concatenation. MHA stands for the standard Multi-head Attention~\cite{vaswani2017attention}.

\noindent \textbf{Divided Fusion.} We provide a modality-specific attention computation manner, \textit{i.e.,} Divided Fusion decomposes the multi-modal fusion into two parallel branches and the final results are summed up with learnable weights.
\begin{equation*}
\mathbf{Y}_{\text{divided}} \!=\! \text{MHA}(\mathbf{\hat{Q}}, \mathbf{H}^k, \mathbf{H}^v) \oplus \text{MHA}(\mathbf{\hat{Q}}, \mathbf{S}^k, \mathbf{S}^v),
\end{equation*}
where $\mathbf{\hat{Q}}$, $\mathbf{H}^k$, $\mathbf{H}^v$, $\mathbf{S}^k$ and $\mathbf{S}^v$  are defined the same as in the joint fusion. $\oplus$ denotes the additive sum with learnable weights.

\noindent \textbf{Hybrid Fusion.} Hybrid Fusion offers a compromise between Joint Fusion and Divided Fusion. Specifically, the query-key multiplication is conducted separately while the query-value multiplication is still in an concatenation format. Suppose there are $n_{h}$ self-attention heads. The query, key and value embeddings are uniformly split into $n_{h}$ segments $\mathbf{\hat{Q}}_{i} \in \mathbb{R}^{N \times d_{h}}$, $\mathbf{H}_{i}^{k}$, $\mathbf{H}_{i}^{v} \in \mathbb{R}^{F \times d_{h}}$, $\mathbf{S}_{i}^{k}$, $\mathbf{S}_{i}^{v}  \in \mathbb{R}^{S \times d_{h}}$, $\{i=1,2, \ldots, n_{h}\}$ along channel dimension, where $d_{h}$ is the dimension of each head and equal to $d/n_{h}$. For each head, we apply hybrid fusion in the form:
\begin{equation*}
\begin{small}
\begin{aligned}
\text{head}_{i}=\left(\operatorname{\sigma}~(\frac{\mathbf{\hat{Q}}_{i} {\mathbf{H}_{i}^{k\top}}}{\sqrt{d_{v}}} ) \otimes \operatorname{\sigma}~(\frac{\mathbf{\hat{Q}}_{i} \mathbf{S}_{i}^{k\top}}{\sqrt{d_{v}}})\right)\left(\mathbf{H}_{i}^{v} \otimes \mathbf{S}^{v}_{i}\right),
\end{aligned}
\end{small}
\end{equation*}
where $\sigma$ is the softmax function. %The query-value correlation computation is conducted separately for each modality while the value branch is simply formed by channel-wise concatenation. 
The outputs of all heads are then again concatenated along the channel dimension and a linear projection is finally applied to produce the final output as follows:
%\begin{equation*}
%\mathbf{Y}_{\text{hybrid}} %=\left(\text{Concat}_{i=1}^{n_h}(\text{head}_i)\right) \mathbf{W}^{O},
%\end{equation*} % $\text{Concat}$ denotes channel-wise concatenation 
\begin{equation*}
\mathbf{Y}_{\text{hybrid}} =\left(\text{head}_0 \otimes \text{head}_1 \otimes \ldots \text{head}_{n_h-1}\right) \mathbf{W}^{O},
\end{equation*}
where $ \mathbf{W}^{O} \in \mathbb{R}^{d \times d}$ is linear projection parameters. %To be a counterpart, we can also set \textit{key} features in concatenation and \textit{value} features in separate. Refer to supplementary material for detail.

\noindent \textbf{Stepwise Fusion.} This fusion manner implements the cross-modality reasoning in a cascaded way, \ie, attention computation is performed between  $\mathbf{\hat{Q}}$ and video features and then propagated to the sentence modality:
\begin{equation*}
\mathbf{Y}_{\text{step}} =  \text{MHA}(\text{MHA}(\mathbf{\hat{Q}}, \mathbf{H}^k, \mathbf{H}^v), \mathbf{S}^k, \mathbf{S}^v).
\end{equation*}
We further discuss more multi-modal fusion mechanisms in Sec.~\ref{sec:4_2} and supplementary materials.

%Considering the modality fusion order, we added an additional variant in the supplementary material which applies language feature attention computation before the vision one. 

%-------------------------------------------------------------------------------------%

\subsection{Many-to-One Matching Loss}\label{sec:3.3}
%-------------------------------------------------------------------------------------%
\textbf{Training}: Based on the Cross-Modality Decoder output, a feed forward network is applied to generate a fixed length predictions $\hat{\mathcal{Y}} = \left\{\hat{y}_{i}\right\}_{i=1}^{N}$, where $\hat{y}_i = (\mathbf{\hat{b}}_i; \hat{c}_i)$ contains temporal segment predictions $\mathbf{\hat{b}}_{i} \in[0,1]^{2}$ and the confidence score $\hat{c}_i \in[0, 1]$. The ground truth is denoted as $\mathcal{Y}$, which contains the segment coordinate $\mathbf{b} \in[0,1]^{2}$. 

GTR applies set prediction loss~\cite{carion2020end,stewart2016end} between the fixed-size output sets and ground-truth. Notably, considering each language query only corresponds to one temporal segment, we adapt the \textit{many-to-many} matching in \cite{carion2020end} to the \textit{many-to-one} version. Specifically, the loss computation is conducted in two consecutive steps. Firstly, we need to determine the optimum prediction slot via the matching cost based on the bounding box similarity and confidence scores as follows:
\begin{equation}
\begin{aligned}
i^* &=\underset{i \in [0, N-1]}{\arg \min } \mathcal{C}_{\mathrm{match}}\left(\hat{\mathcal{Y}}, \mathcal{Y}\right) \\
&= \underset{i \in [0, N-1]}{\arg \min }\left[-\hat{c}_{i}+\mathcal{C}_{\mathrm{box}}(\mathbf{b}, \mathbf{\hat{b}}_{i})\right].
\label{eq:match}
\end{aligned}
\end{equation}
In our many-to-one matching loss, the optimal match requires only one iteration of $N$ generated results, rather than checking all possible permutations as in \cite{carion2020end}, which greatly simplifies the matching process. For $\mathcal{C}_{\mathrm{box}}(\cdot, \cdot)$, we define $\mathcal{C}_{\mathrm{box}}=\lambda_{\ell_{1}}\|b-\hat{b}_{i}\|_{1}+\lambda_{\mathrm{iou}} \mathcal{C}_{\mathrm{iou}}(b, \hat{b}_{i})$ with weighting parameters $\lambda_{\ell_{1}}$, $\lambda_{\mathrm{iou}} \in \mathbb{R}$. Here $\mathcal{C}_{\mathrm{iou}}(\cdot, \cdot)$ is a scale-invariant generalized intersection over union in \cite{rezatofighi2019generalized}. 
Then the second step is to compute the loss function between the matched pair:
\begin{equation}
\mathcal{L}_{\text{set}}(y, \hat{y}) = -\log \hat{c}_{i^*}+\mathcal{C}_{\mathrm{box}}(\mathbf{b}, \mathbf{\hat{b}}_{i^*}),
\label{eq:loss}
\end{equation}
where $i^*$ is the optimal match computed in Eq.~\eqref{eq:match}.
\noindent \textbf{Inference}: During inference, the predicted segment set is generated in one forward pass. Then the result with the highest confidence score is selected as the final prediction. The whole inference process requires no predefined threshold values or specific post-processing processes.

\section{Experiments}\label{expes}
We first introduce experimental settings in Sec.~\ref{sec:4_1}. Then, we present detailed exploratory studies on the design of GTR in Sec.~\ref{sec:4_2}. The comparisons with SOTA methods are discussed in Sec.~\ref{sec:4_3}, and we show visualization results in Sec.~\ref{sec:4_4}. More results are left in supplementary materials.
%-------------------------------------------------------------------------------------%
%-------------------------------------------------------------------------------------%
\subsection{Settings}\label{sec:4_1}
%-------------------------------------------------------------------------------------%
%------------------------------------------------------------------
%\begin{wraptable}{r}{0.75\textwidth}
\begin{table*}[t]
\centering
\vspace{-5mm}
%-----------------------------------------------------------------------------
%\subfloat[{Spatial configuration}.\label{tab:3:1}]{
\subfloat[{\footnotesize{Spatial Configuration}}]{
\tablestyle{3pt}{1.1}
\small{
\begin{tabular}{l|x{25}x{25}x{25}}
\multicolumn{1}{l|}{Models}& $R_1^{0.5}$ & $R_1^{0.7}$  & {GFLOPs}\\
\shline
%GTR-B/4   & 50.25 & 28.74 & 25.50 \\ 
%\rowcolor{green!30} 
GTR-B/8   & \textbf{49.67} & \textbf{28.45} & \textbf{20.55} \\
GTR-B/12  & 44.32 & 24.62 & 8.71 \\ 
GTR-B/16  & 43.14 & 23.93 & 5.44 \\ 
GTR-B/24  & 42.01 & 23.82 & 1.96 \\
\multicolumn{4}{c}{~}\\ % A blank line
\end{tabular}}
}%}
%------------------------------------------------------------------------------
\subfloat[{\footnotesize{Temporal Configuration}}]{
\tablestyle{3pt}{1.1}
\small{
\begin{tabular}{c|x{25}x{25}x{25}}
	\multicolumn{1}{c|}{$k_t$}& $R_1^{0.5}$ & $R_1^{0.7}$  & {GFLOPs}\\
	\shline
	%1* & 46.10 & 26.27 & 46.36 \\
	%\rowcolor{green!30} 
	3 & \textbf{49.67} & \textbf{28.45} & \textbf{20.55} \\ 
	5 & 47.34 & 26.12 & 17.83 \\ 
	7 & 47.87 & 26.91 & 15.03 \\ 
	9 & 48.04 & 27.02 & 13.91 \\ 
    \multicolumn{4}{c}{~}\\ % A blank line
\end{tabular}}
}
%------------------------------------------------------------------------------
\subfloat[{\footnotesize{Overlapping Exploration}}]{
	\tablestyle{3pt}{1.1}
\footnotesize{
%\begin{tabular}{r|r|ccccc}
\begin{tabular}{c|x{40}|x{25}x{25}x{25}}
	Models &\multicolumn{1}{c|}{Stride}& $R_1^{0.5}$ & $R_1^{0.7}$  & {GFLOPs}\\
	\shline
    %\rowcolor{green!30} 
    \multirow{2}{*}{Temporal} & (8, 8, 1) & 49.73 & 28.51 & 41.05 \\
    %~ & \cellcolor{green!30}{(8, 8, 2)} & \cellcolor{green!30}{48.36} & \cellcolor{green!30}{27.21} & \cellcolor{green!30}{20.55} \\
    ~ & {(8, 8, 2)} & \textbf{49.67} & \textbf{28.45} & \textbf{20.55} \\
    \hline
	\multirow{2}{*}{Spatial} & (4, 4, 3) & 44.54 & 23.12 & 54.51 \\ 
	~ & (6, 6, 3) & 44.23 & 22.81 & 24.26 \\ 
	\hline
	None & (8, 8, 3) & 43.73 & 22.45 & 14.69 \\ 
	 %\multicolumn{4}{c}{~}\\ % A blank line
\end{tabular}}
}
\vspace{-3mm}
\caption{\footnotesize Input ablations for the GTR-B model\footref{GTRB} on ANet dataset (\%).}
\vspace{-3mm}
\label{tab:InputAbla}
\end{table*}
%\end{wraptable}
%-------------------------------------------------------------------------------------%
%-------------------------------------------------------------------------------------%
\textbf{Datasets.} We evaluated our GTR on three challenging video grounding benchmarks: 1) \textbf{ActivityNet Captions (ANet)}~\cite{krishna2017dense}: %It contains 20,000 untrimmed videos with 100,000 descriptions. 
The average video length is around 2 minutes, and the average length of ground-truth video moments is 40 seconds. By convention, 37,417 video-query pairs for training, 17,505 pairs for validation, and 17,031 pairs for testing. 2) \textbf{Charades-STA}~\cite{gao2017tall}: %It is re-labeled by Gao \textit{et al.} based on the Charades~\cite{sigurdsson2016hollywood}. 
The average length of each video is around 30 seconds. Following the official splits, 12,408 video-query pairs for training, and 3,720 pairs for testing. 3) \textbf{TACoS}~\cite{regneri2013grounding}: It is a challenging dataset focusing on cooking scenarios. Following previous works~\cite{gao2017tall}, we used 10,146 video-query pairs for training, 4,589 pairs for validation, and 4,083 pairs for testing.

%-------------------------------------------------------------------------------------%
\noindent \textbf{Evaluation Metrics.} Following prior works, we adopt “R@n, IoU@m” (denoted as $R^m_n$) as the metrics. Specifically, $R^m_n$ is defined as the percentage of at least one of top-n retrieved moments having IoU with the ground-truth moment larger than m.
%-------------------------------------------------------------------------------------%

\noindent \textbf{Modal Variants.} Following the practice of Visual Transformers or BERTs~\cite{dosovitskiy2020image,devlin2019bert}, we also evaluate three typical model sizes: GTR-Base (\textbf{GTR-B}, $N_v=4$, $N_s=4$, $N_d=6$, $d=320$), GTR-Large (\textbf{GTR-L}, $N_v=6$, $N_s=6$, $N_d=8$, $d=320$), and GTR-Huge (\textbf{GTR-H}, $N_v=8$, $N_s=8$, $N_d=8$, $d=512$).\footref{supple}
%-------------------------------------------------------------------------------------%

\noindent \textbf{Implementation Details.} For input video, the sampling rate $1/\gamma_{\tau}$ was set to $1/8$, all the frames were resized to $112 \times 112$, the kernel shape and stride size were set to (8, 8, 3) and (8, 8, 2), respectively. We used AdamW~\cite{loshchilov2017decoupled} with momentum of 0.9 as the optimizer. The initial learning rate and weight decay were set to $10^{-4}$. All weights of the encoders and decoders were initialized with Xavier init, and the cubic embedding layer was initialized from the ImageNet-pretrained ViT~\cite{dosovitskiy2020image}. We used random flip, random crop, and color jitter for video data augmentation. Experiments were conducted on 16 V100 GPUs with batch size 64. % The training process lasted for 10 epochs.

%-------------------------------------------------------------------------------------%
%\subsection{Ablation Study}\label{sec:4_2}
\subsection{Empirical Studies and Observations}\label{sec:4_2}

%-----------------------------------------------------------------------
%\begin{wraptable}{r}{0.5\textwidth}
\begin{table}
\vspace{-2mm}
\small
\centering
\renewcommand\arraystretch{1.1}
\setlength{\tabcolsep}{3pt}{
\begin{tabular}{x{50}|x{35}x{35}x{35}}
Models & $R_1^{0.5}$  & $R_1^{0.7}$ & GFLOPs \\
\shline
Framewise & 46.10 & 26.27 & 46.36 \\
Cubic & \textbf{49.67} & \textbf{28.45} & \textbf{20.55} \\
\end{tabular}
}
\vspace{-3mm}
\caption{\footnotesize {Cubic \emph{vs.} Framewise Embedding on ANet(\%).}}
\label{tab:vsViT}
\vspace{-3mm}
\end{table}
%\end{wraptable}
%-------------------------------------------------------------------------------------%
%-------------------------------------------------------------------------------------%
%-----------------------------------------------------------------------
\begin{comment}
%\begin{wraptable}{r}{0.5\textwidth}
\begin{table}
%\vspace{-12mm}
\small
\centering
\renewcommand\arraystretch{1.1}
\setlength{\tabcolsep}{3pt}{
\begin{tabular}{r|ccccc}
%\begin{tabular}{r|x{26}x{26}x{28}x{30}}
	\multicolumn{1}{r|}{Method}& $R_1^{0.3}$ & $R_1^{0.5}$  & Param(M) & GFLOPs\\
	\shline
	Early Fusion & 28.25 & 19.56  & 12.86 & 11.31  \\ 
	Conditional Fusion & 24.62 & 14.37 & 12.42 & 10.60  \\
	\hline
	Joint Fusion & 33.69 & 23.72 & 16.62 & 13.91 \\ 
	Divided Fusion & 37.82 & 26.31  & 25.16 & 21.02 \\
	Hybrid Fusion & \textcolor{blue}{\textbf{37.01}} & \textcolor{blue}{\textbf{25.77}} & \textcolor{blue}{\textbf{20.94}} & \textcolor{blue}{\textbf{17.25}} \\
	Stepwise Fusion & \textcolor{red}{\textbf{39.34}} & \textcolor{red}{\textbf{28.34}} & \textcolor{red}{\textbf{25.98}} & \textcolor{red}{\textbf{20.55}} \\
\end{tabular}}
\vspace{-2mm}
\caption{\footnotesize Multi-modal fusion comparisons on TACoS.(\%).}
\label{tab:fuseAbla}
\vspace{-2mm}
\end{table}
%\end{wraptable}
\end{comment}
%-------------------------------------------------------------------------------------%
%-----------------------------------------------------------------------
%\begin{wraptable}{r}{0.5\textwidth}
\begin{table}
\small
\centering
\renewcommand\arraystretch{1.1}
\setlength{\tabcolsep}{3pt}{
    \begin{tabular}{r|ccccc}
    %\begin{tabular}{r|x{26}x{26}x{28}x{30}}
    	\multicolumn{1}{r|}{Models}& $R_1^{0.5}$ & $R_1^{0.7}$  & Param(M) & GFLOPs\\
    	\shline
    	Early Fusion & 39.35 & 19.64  & 12.86 & 11.31  \\ 
    	Conditional Fusion & 35.25 & 14.57 & 12.42 & 10.60  \\
    	\hline
    	Joint Fusion & 42.51 & 21.53 & 16.62 & 13.91 \\ 
    	Divided Fusion & 46.91 & 26.21  & 25.16 & 21.02 \\
    	Hybrid Fusion & \textcolor{blue}{\textbf{46.82}} & \textcolor{blue}{\textbf{25.45}} & \textcolor{blue}{\textbf{20.94}} & \textcolor{blue}{\textbf{17.25}} \\
    	Stepwise Fusion & \textcolor{red}{\textbf{49.67}} & \textcolor{red}{\textbf{28.45}} & \textcolor{red}{\textbf{25.98}} & \textcolor{red}{\textbf{20.55}} \\
    \end{tabular}}
\vspace{-2mm}
\caption{\footnotesize Multi-modal fusion comparisons on ANet.(\%).}
\label{tab:fuseAbla}
\vspace{-2mm}
\end{table}
%\end{wraptable}
%-------------------------------------------------------------------------------------%
%-------------------------------------------------------------------------------------%

%-------------------------------------------------------------------------------------%

%-------------------------------------------------------------------------------------
%\newcommand{\tabincell}[2]{\begin{tabular}{@{}#1@{}}#2\end{tabular}}
%-------------------------------------------------------------------------------------%
%-----------------------------------------------------------------------
%\begin{wraptable}{r}{0.6\textwidth}
\begin{table*}[ht]
	\vspace{-2mm}
	\small
	\centering
	\renewcommand\arraystretch{1.1}
%	\scriptsize{
	\setlength{\tabcolsep}{3.5pt}{
		\resizebox{\linewidth}{!}{
	\begin{tabular}{r|cccc|ccccc|ccccc|c}
	\multirow{2}{*}{Models} & \multicolumn{4}{c|}{ActivityNet Captions} & \multicolumn{5}{c|}{Charades-STA} & \multicolumn{5}{c|}{TACoS} & \multirow{2}{*}{\small{Param}}\\
	~ & $R_1^{0.5}$ & $R_1^{0.7}$  & $R_5^{0.5}$ & $R_5^{0.7}$ & Feat. & $R_1^{0.5}$ & $R_1^{0.7}$  & $R_5^{0.5}$ & $R_5^{0.7}$  & $R_1^{0.3}$ & $R_1^{0.5}$  & $R_5^{0.3}$ & $R_5^{0.5}$ & QPS & ~ \\
	\shline
	2D-TAN & 44.51 & 26.54 & 77.13 & 61.96 & VGG & 39.81 & 23.25 & 79.33 & 52.15 & 37.29 & 25.32 & 57.81  & 45.04 & 2.36 & 93.37\\ 
	DRN & 45.45 & 24.36 & 77.97 & 50.30 & I3D & 53.09 & 31.75 & 89.06 & 60.05 & --- & 23.17 & --- & 33.36 & 1.82 & 108.34\\
	SCDM & 36.75 & 19.86  & 64.99 & 41.53 & I3D & 54.44 & 33.43 & 74.43 & 58.08 & 26.11 & 21.17 & 40.16  & 32.18 & 1.28 & 93.82\\ 
	QSPN  & 33.26 & 13.43  & 62.39 & 40.78 & C3D & 35.60 & 15.80  & 79.40  & 45.40 & 20.15& 15.23 &36.72 &25.30 & 1.14 & 95.03\\
	CBP & 35.76 & 17.80 & 65.89 & 46.20 & C3D & 36.80 & 18.87& 70.94 &50.19& 27.31& 24.79 &43.64& 37.40 & 1.92 & 98.55\\
	%LGI & 41.51 & 23.07 & - & - & I3D & 59.46 & 35.48 & - & -
	\hline
	2D-TAN* & 45.21 & 27.35 & 77.60 & 62.32 & VGG  &40.32  &23.63 & 80.22 & 52.57  &37.89  &25.99 & 58.23  &45.21 & 2.36 & 93.37\\
	DRN* & 46.34 & 24.92 & 78.12 & 50.93 & I3D & 53.82 & 32.34 & 89.74 & 60.23 & --- & 23.84 & --- &33.91 & 1.82 & 108.34 \\
	\hline
	GTR-B (Ours) & {\color{blue}\textbf{49.67}} & {\color{blue}\textbf{28.45}} & {\color{blue}\textbf{79.83}} & {\color{blue}\textbf{64.34}} & --- & \color{blue}\textbf{62.45} & \color{blue}\textbf{39.23}&  \color{blue}\textbf{91.40} & \color{blue}\textbf{61.76} & \color{blue}\textbf{39.34} & \color{blue}\textbf{28.34} & \color{blue}\textbf{60.85}&  \color{blue}\textbf{46.67} & \color{blue}\textbf{13.4} & \color{blue}\textbf{25.98}\\
	GTR-L (Ours) & {\textbf{50.43}} & {\textbf{28.91}} & {\textbf{80.22}} & {\textbf{64.95}} & --- & \textbf{62.51} & \textbf{39.56}&  \textbf{91.62} & \textbf{61.97} & \textbf{39.93} & \textbf{29.21} & \textbf{61.22}&  \textbf{47.10} & \textbf{12.7}  & \textbf{40.56} \\
	GTR-H (Ours) & {\color{red}\textbf{50.57}} & {\color{red}\textbf{29.11}} & {\color{red}\textbf{80.43}} & {\color{red}\textbf{65.14}} & --- & \color{red}\textbf{62.58} & \color{red}\textbf{39.68}&  \color{red}\textbf{91.62} & \color{red}\textbf{62.03} & \color{red}\textbf{40.39} & \color{red}\textbf{30.22} & \color{red}\textbf{61.94} & \color{red}\textbf{47.73} & \color{red}\textbf{11.8}  & \color{red}\textbf{61.35} \\
	\end{tabular}}}%}
	\vspace{-2.5mm}
	\caption{\footnotesize Performance comparisons on three benchmarks(\%). All the reported results on ANet and TACoS datasets are based on C3D~\cite{tran2015learning} extracted feature. ``*'' denotes finetuning on corresponding backbones. For pair comparisons, Parma (M) includes the parameter of feature extractor (C3D). {\color{blue}\textbf{GTR-B}} is more efficient while {\color{red}\textbf{GTR-H}} achieves the highest recall.}
	\label{tab:compSOTA}
	\vspace{-2mm}
\end{table*}
%\end{wraptable} 
%-------------------------------------------------------------------------------------%

%-------------------------------------------------------------------------------------

%------------------------------------------------------\-------------------------------
In this subsection, we conducted extensive studies on different design choices of GTR, and tried to answer four general questions: \textbf{Q1:} How to transform a raw video into visual tokens? \textbf{Q2:} How to fuse the video and text features? \textbf{Q3:} Are there any design principles to make a stronger Transformer decoder? \textbf{Q4:} Are there any good training recipes?

% What is the optimal input paradigm?
% What makes an effective transformer?
% What are the good training recipes?
%\textcolor{red}{Due to page limit, we only showed ablation studies on the ActivityNet Caption dataset and results on Charades-STA are given in Appendix X.}

%-------------------------------------------------------------------------------------%
\subsubsection{Visual Tokens Acquisition (Q1)}
\noindent\textbf{Settings.} In cubic embedding layer, there are two sets of hyper-parameters (kernel shape $(k_w, k_h, k_t)$ and stride size $(s_w, s_h, s_t)$). We started our studies from a basic GTR-B model\footnote{The baseline GTR-B is with the stepwise fusion strategy, and with $s_w=k_w=s_h=k_h=8$, $k_t=3$, $s_t=2$.\label{GTRB}}, and discussed design choices from three aspects: 1) \emph{Spatial configuration.} We compared four GTR-B variants with different kernel spatial size $k_w,k_h=\{8,12,16,24\}$, and denoted these models as GTR-B/*. Results are reported in Table~\ref{tab:InputAbla} (a). 2) \emph{Temporal configuration.} We compared four GTR-B variants with different kernel temporal depths $k_t = \{3,5,7,9\}$. Results are reported in Table~\ref{tab:InputAbla} (b). 3) \emph{Overlapping exploration.} We conducted experiments with several different stride sizes $(s_w, s_h, s_t)$ in Table~\ref{tab:InputAbla} (c), which corresponds to three basic types (temporal overlapping, spatial overlapping, and non-overlapping). 

\noindent\textbf{Observations.} 1) Models with smaller patch size (\eg, GTR-B/8) achieve better performance yet at the cost of the dramatic increase of FLOPs. 2) Models with larger kernel temporal depth ($k_t$) will not always achieve better results. It is worth noting that our cubic embedding will degrade into the prevalent framewise partition in vision Transformers by setting $k_t=s_t=1$. We further compared cubic embedding with this special case in Table~\ref{tab:vsViT}, and the results show the superiority of our cubic embedding layer. 3) Compared to non-overlapping and spatial overlapping, temporal overlapping can help to improve model performance significantly. Besides, the performance is not sensitive to the \emph{overlapping degree} in all overlapping cases. 

% 4) We add comparisons when using framewise partition ($s_t\!=1$) in ViT to embed video data. The results on ActivityNet Caption (Table.~\ref{tab:vsViT}) demonstrate the superiority of our cubic embedding layer.
% For instance, stride (8, 8, 2) shares similar performance with stride (8, 8, 1) yet the GFLOPs is much lower.
%-------------------------------------------------------------------------------------%
%\input{tables_figs/tabWithViT} % Move before
%-------------------------------------------------------------------------------------%

\noindent\textbf{Guides.} \emph{The temporal overlapping sampling strategy can greatly boost the performance of the Cubic Embedding layer at an affordable overhead.}

%-------------------------------------------------------------------------------------%
%\begin{figure}[t]
%\centering
%\includegraphics[width=.23 \textwidth]{figs/vsCrop.pdf} \\
%\includegraphics[width=.23 \textwidth]{figs/vsRate.pdf}
%\caption{\small Left: FLOPs \textit{v.s.} Frame Crop Size; Right: FLOPs \textit{v.s.} Frame Sample Rate.}
%\label{fig:vsCropvsRate}
%\end{figure}
\begin{figure}[t]
\centering
\includegraphics[width=0.48\textwidth]{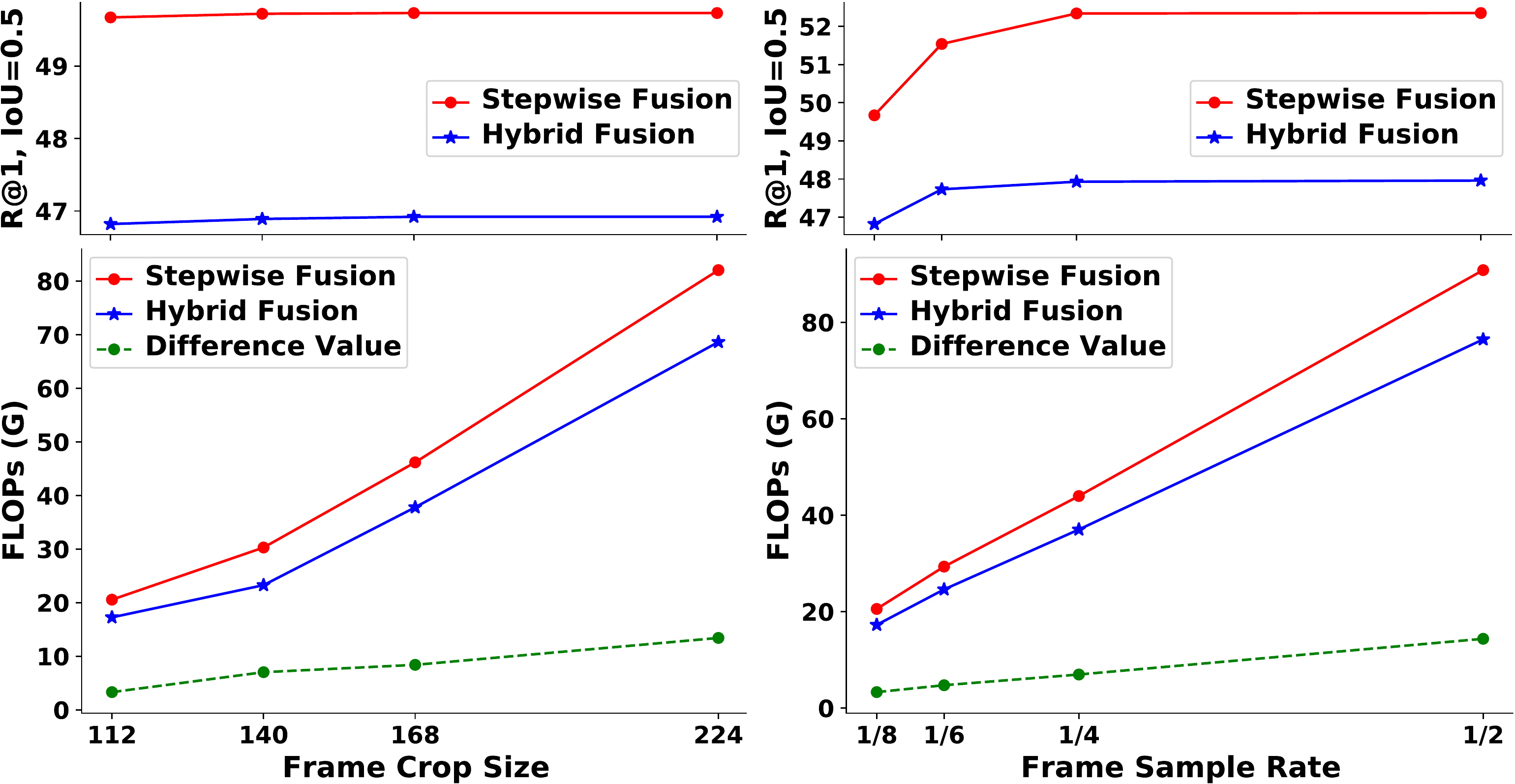}
%\vspace{-20pt}
\vspace{-6mm}
%\caption{\footnotesize{\textbf{Left:} FLOPs \textit{v.s.} Frame Crop Size; \textbf{Right:} FLOPs \textit{v.s.} Frame Sample Rate. when frame crop size or sampling rate increases, The FLOPS values of both methods increased, but the difference does not increase significantly.}}
%\caption{\footnotesize{\textbf{Left:} $R_1^{0.5}$ \& FLOPs \textit{v.s.} Frame Crop Size; \textbf{Right:} $R_1^{0.5}$ \& FLOPs \textit{v.s.} Frame Sample Rate.}}
\caption{\footnotesize{\textbf{Top:} $R_1^{0.5}$ \textit{v.s.} Frame Crop Size \& Sample Rate; \textbf{Bottom:} FLOPs \textit{v.s.} Frame Crop Size \& Sample Rate.}}
\vspace{-3mm}
\label{fig:vsCropvsRate}
\end{figure}

%-------------------------------------------------------------------------------------%
\subsubsection{Multi-modal Fusion Mechanisms (Q2)}

\noindent\textbf{Settings.} As mentioned in Sec.~\ref{sec:3.2}, we design four types of multi-modal fusion mechanism in the decoder (\ie, joint/divided/hybrid/stepwise fusion), and we group all these fusion mechanisms as \emph{late fusion}. Meanwhile, we also propose two additional fusion mechanisms\footref{supple}: 1) \emph{Early fusion}: The multi-modal features are fused before being fed into the decoder. 2) \emph{Conditional fusion}: The language features act as conditional signals of segment queries of the decoder. Results are reported in Table~\ref{tab:fuseAbla}.
% \noindent \textbf{A2}: To explore the best way of multi-modality fusion, in Table~\ref{tab:fuseAbla}, we present the results for the proposed four multi-modal fusion schemes and two additional \textit{early fusion} and \textit{conditional fusion} methods on TACoS dataset. In early fusion, video and language hidden states are concatenated before fed to a standard decoder; In conditional fusion, language features act as a conditional information and are concatenated on the segment queries. See appendix for details. In contrast, the above four fusion methods can be collectively referred to as \textit{late fusion} since they all conduct modality fusion within the decoder.

\noindent\textbf{Observations.} 1) All four late fusion models outperform the early fusion and conditional fusion ones. 2) Among late fusion models, the stepwise fusion model achieves the best performance by using the largest number of parameters. 3) The performance is not sensitive to the crop size, but is positively correlated with sample rate and converges gradually(cf. Figure~\ref{fig:vsCropvsRate}). 4) We find that the FLOPs difference does not increase significantly regarding the crop size and sample rate (cf. Figure~\ref{fig:vsCropvsRate}).

\noindent\textbf{Guides.} \emph{Stepwise fusion is the optimal fusion mechanism, even for long and high-resolution videos.}

\subsubsection{Decoder Design Principles (Q3)}
\noindent\textbf{Settings.} To explore the key design principles of a stronger multi-modal Transformer decoder, we considered two aspects: 1) \emph{Deeper vs. Wider}, \ie, whether the decoder should go deeper or wider? Based on the basic GTR-B model\footref{GTRB}, we designed two GTR-B variants with nearly equivalent parameters: GTR-B-Wide ($d=352$) and GTR-B-Deep ($N_d=8$)\footref{supple}. The results are reported in Table~\ref{tab:Q3} (a). 2) \emph{Columnar vs. Shrink vs. Expand.}, \ie, how to design the attention head number in different layers. We tried three different choices: i) the same number of heads in all layers (columnar); ii) gradually decrease the number of heads (shrink); iii) gradually increase the number of heads (expand). Results are reported in Table~\ref{tab:Q3} (b).
% GTR-B-Wide with the hidden dimension extended to 256 and GTR-B-Deep with layer number increased to 8. 

\noindent\textbf{Observations.} 1) Under the constraint of similar parameters, the deep model (GTR-B-Deep) outperforms the wide counterpart (GTR-B-Wide). 2) The model with the expanding strategy (GTR-B-Expand) achieves the best performance while the shrinking one (GTR-B-Shrink) is the worst.

\noindent\textbf{Guides.} \emph{Going deeper is more effective than going wider and progressively expanding the attention heads leads to better performance.} 
% Interestingly, it is consistent with the ResNet designing rule (\ie, enlarging feature channels with the growth of network depth).

%-------------------------------------------------------------------------------------%
%\input{tables_figs/tabDeepWide} % Merge to tabQ3
%-------------------------------------------------------------------------------------%
%-------------------------------------------------------------------------------------%
%\input{tables_figs/tabExpand} % Merge to tabQ3
%-------------------------------------------------------------------------------------%
%------------------------------------------------------------------
%\begin{wraptable}{r}{0.75\textwidth}
\begin{table}[t]
\centering
\subfloat[{\footnotesize{Wide \textit{vs.} Deep.}}]{
\tablestyle{3pt}{1.1}
\small{
\begin{tabular}{l|c|cc|cc}
	\multirow{2}{*}{Models} & \multirow{2}{*}{Param(M)} & \multicolumn{2}{c|}{ActivityNet} & \multicolumn{2}{c}{TACoS} \\
	~&  ~&  $R^{0.5}_1$ & $R^{0.7}_1$ &  $R^{0.3}_1$ & $R^{0.5}_1$ \\
	\shline
	{GTR-B-Wide} & 34.05 & 49.94 & 28.62 & 39.52 & 28.70  \\ 
	{GTR-B-Deep} & 33.89 & \textbf{50.15} & \textbf{28.74} & \textbf{39.60} & \textbf{29.04} \\
	%\multicolumn{8}{c}{~}\\ % A blank line
	%\hline
	%{GTR-L-Wide} & 52.25 & 50.51 & 28.99 & 62.54  & 39.60 & 40.12 & 29.98 \\
	%{GTR-H} & 52.46 & 50.57 & 29.11 & 62.58  & 39.68 & 40.39 & 30.22 \\ 
\end{tabular}
}}%}%}
\\
\vspace{-3mm}
%------------------------------------------------------------------------------
\subfloat[{\footnotesize{Columnar \textit{vs.} Shrink \textit{vs.} Expand on ANet.}}]{
\tablestyle{3pt}{1.1}
\small{
%\resizebox{0.8\linewidth}{!}{
\begin{tabular}{l|c|cc}
	{Models} & {Heads} &  $R^{0.3}_1$ & $R^{0.5}_1$  \\
	\shline
	{GTR-B-Columnar} & [4, 4, 4, 4, 4, 4] & 49.42 & 28.34  \\ 
	{GTR-B-Shrink} & [8, 8, 4, 4, 2, 1]  & 49.01 & 27.95 \\
	{GTR-B-Expand} & [1, 2, 4, 4, 8, 8]  & \textbf{49.67} & \textbf{28.45} \\
	%\multicolumn{4}{c}{~}\\ % A blank line
	%\multicolumn{4}{c}{~}\\ % A blank line
	\end{tabular}
}}
%}
%\\
%------------------------------------------------------------------------------
\vspace{-2mm}
\caption{\footnotesize{\textbf{(a)} Performance (\%) comparisons between the \emph{wide} variant and \emph{deep} variant. \textbf{(b)} Performance (\%) comparisons between different attention head number choices.}}
\vspace{-3mm}
\label{tab:Q3}
\end{table}
%\end{wraptable}
%\input{tables_figs/tabQ4}
%-------------------------------------------------------------------------------------%
\subsubsection{Training Recipes (Q4)}
%-------------------------------------------------------------------------------------%
\begin{figure*}[htbp]
    \begin{minipage}[c]{0.35\linewidth}
        \captionsetup{type=table} %% tell latex to change to table
        \scalebox{0.75}{
            %%%%%%%%%%%%%%%%%%%%%%%% efficiency %%%%%%%%%%%%%%%%%%%%
            \begin{tabular}{l|cccc}
        	\multicolumn{1}{l|}{Models}& $R_1^{0.5}$ & $R_1^{0.7}$ \\
        	\shline
        	Baseline & 41.50 & 20.13  \\ 
        	%\rowcolor{green!30} 
        	~+ Pretrained Weight & \textbf{47.12}$_{\color{red}{+5.62}}$ & \textbf{26.03}$_{\color{red}{+5.90}}$ \\
        	~+ Random flip & 47.59$_{+0.47}$ & 26.69$_{+0.66}$ \\
        	~+ Random crop & 47.61$_{+0.02}$ & 26.81$_{+0.12}$ \\
        	%\rowcolor{green!30} 
        	~+ Color Jitter & \textbf{49.67}$_{\color{red}{+2.06}}$ & \textbf{28.45}$_{\color{red}{+1.64}}$ \\ %$_{\color{red}{+1.64}}$ 
        	 %\multicolumn{3}{c}{~}\\ % A blank line
            \end{tabular}
        }
        \vspace{-2mm}
        \caption{Results of different training tricks of the baseline GTR-B model\footref{GTRB} on the ANet dataset.}
        \label{tab:receipt}
    \end{minipage} 
    \hfill
    \begin{minipage}[l]{0.6\linewidth}
          \includegraphics[width=0.95\linewidth]{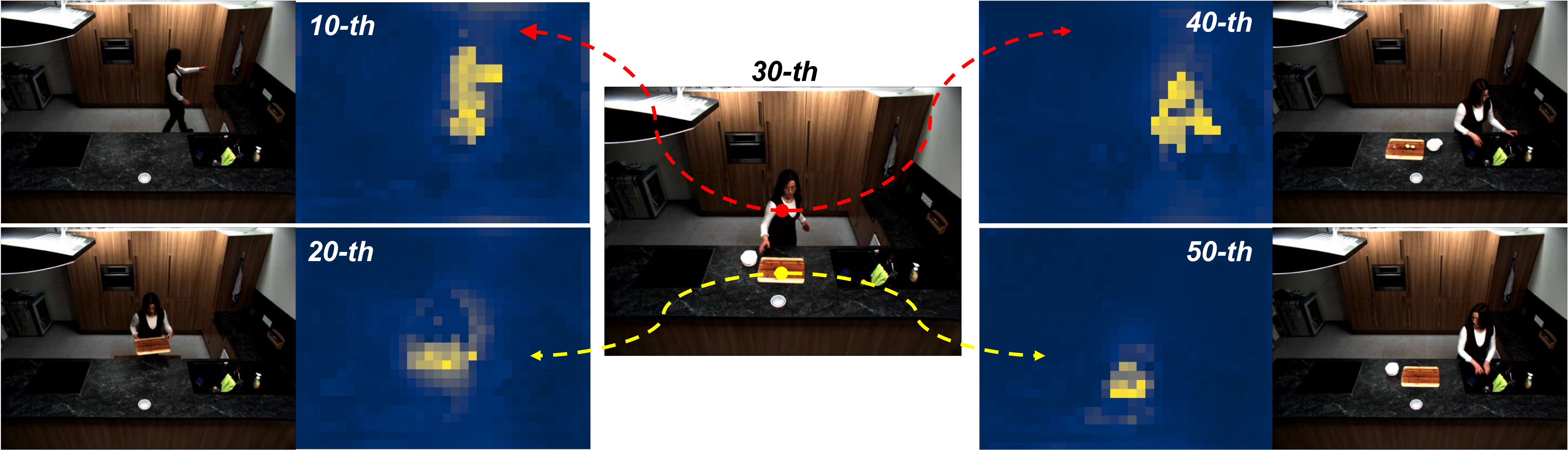}
          \vspace{-3mm}
          \caption{A visualization example of the self-attention in the video encoder.}
          \label{fig:vidEncoderAtten}
           \vspace{-2mm}
    \end{minipage}
\end{figure*}
%-------------------------------------------------------------------------------------%
\noindent\textbf{Settings.} Due to the lack of inductive biases, visual Transformers always over-rely on large-scale datasets for training~\cite{dosovitskiy2020image,touvron2020training}. To make multi-modal Transformers work on relatively small multi-modal datasets, we discussed several common training tricks: 1) \emph{Pretrained weights.} Following the idea of pretrain-then-finetune paradigm in CNN-based models\footnote{Modern CNN models always rely on pretrained weights from large-scale datasets as initialization (\eg, ImageNet), and then finetune the weights on specific downstream tasks.}, we initialize our video cubic embedding layer from the pretrained ViT. Specifically, we initialized our 3D linear projection filters by replicating the 2D filters along the temporal dimension. 2) \emph{Data augmentations.} To study the impact of different data augmentation strategies, we apply three typical choices sequentially, \ie, random flip, random crop, and color jitter. Results are reported in Table~\ref{tab:receipt}.

\noindent\textbf{Observations.} 1) Using the pretrained cubic embedding weights can help to improve model performance significantly. 2) Color jitter brings the most performance gains among all visual data augmentation strategies. We conjecture that this may be due to the fact that color jitter can change the visual appearance without changing the structured information, which is critical for multi-modal tasks.
% Interestingly, our finding is consistent with \cite{chen2020simple}. 

\noindent\textbf{Guides.} \emph{Using pretrained embedding weights and color jitter video augmentation are two important training tricks for multi-modal Transformers.}
%-------------------------------------------------------------------------------------%
\begin{figure}[t]
	\centering
%	\begin{subfigure}[b]{0.35\textwidth}
		\centering
		%%\includegraphics[width=0.3\textwidth]{figs/wordCrossAtten.pdf}
		%\caption{}
%	\end{subfigure} 
%%\quad
%	\begin{subfigure}[b]{0.46\textwidth}
		%%\centering
		%\includegraphics[width=0.42\textwidth]{figs/videoDecoderCompress.pdf}
		\includegraphics[width=0.4\textwidth]{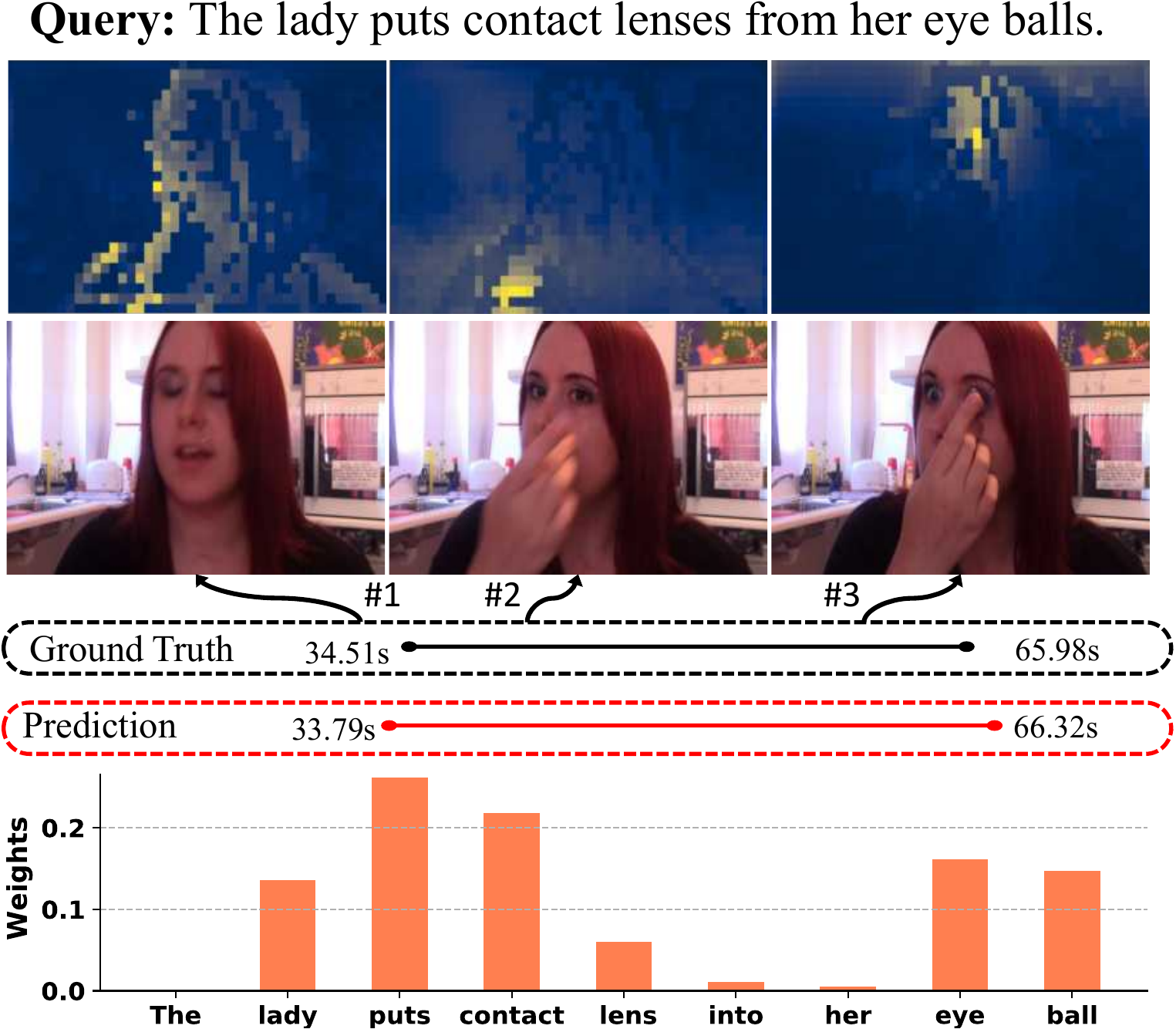} 
		%\includegraphics[width=0.45\textwidth]{figs/decoderAtten.pdf} 
		%\caption{}
%	\end{subfigure}
    \vspace{-8pt}
	\caption{\footnotesize {\textbf{Top:} A visualization example of the cross-attention in the decoder. \textbf{Bottom:} Visualization of the sentence attention weights.}}
	\vspace{-5mm}
	\label{fig:DecoderAtten}
\end{figure}
\subsection{Comparisons with State-of-the-Arts} \label{sec:4_3}
%-------------------------------------------------------------------------------------%
%\input{tables_figs/tabSOTA} % Move Before
%\input{tables_figs/figVidEncoderAtten} 
%-------------------------------------------------------------------------------------%
%\input{tables_figs/figDecoderAtten}
%-------------------------------------------------------------------------------------%
\noindent\textbf{Settings.} We compared three variants of GTR (\ie, GTR-B, GTR-L, and GTR-H) to state-of-the-art video grounding models:  \textbf{2D-TAN}~\cite{zhang2020learning}, \textbf{DRN}~\cite{zeng2020dense}, \textbf{SCDM}~\cite{yuan2019semantic}, \textbf{QSPN}~\cite{xu2019multilevel}, \textbf{CBP}~\cite{wang2020temporally}. These video grounding methods were based on pre-extracted video features (\eg, C3D or I3D) while our GTRs were trained in an end-to-end manner. For more fair comparisons, we selected two best performers (2D-TAN and DRN), and re-trained them by finetuning the feature extraction network. Results are reported in Table~\ref{tab:compSOTA}\footref{supple}.

\noindent\textbf{Results.} From Table~\ref{tab:compSOTA}, we have the following observations: 1) All three GTR variants outperform existing state-of-the-art methods on all benchmarks and evaluation metrics. Particularly, the GTR-H achieves significant performance gains on more strict metrics (\eg, 6.25\% and 4.23\% absolute performance differences on Charades-STA with metric $R^{0.7}_1$ and TACoS with metric $R^{0.5}_1$, respectively.) 2) The inference speed of all GTR variants are much faster than existing methods (\eg, 13.4 QPS in GTR-B vs. 2.36 QPS in 2D-TAN). Meanwhile, our GTR has fewer parameters than existing methods (\eg, 25.98M in GTR-B vs. 93.37M in 2D-TAN).

% On ActivityNet Captions, we observe that our method consistently outperforms the state-of-the-art methods under all IoU thresholds. Under the strictest matric R@1,IoU@0.7, our method reaches 28.45\%. On Charades-STA, our method also shows superior performance, \ie, it surpasses the previous best model (CBLN) with 1.2\% absolute improvements in the R@1,IoU@0.7 metrics. The consistent superior results on both datasets signify the effectiveness of our proposed method. Exhilaratingly, on the challenging TACoS dataset, our GTR-B also achieves quite competitive results with R@1, IoU@0.7 reaching 28.34\%.

%-------------------------------------------------------------------------------------%
%-------------------------------------------------------------------------------------%
\subsection{Visualizations} \label{sec:4_4}
%-------------------------------------------------------------------------------------%

\noindent\textbf{Self-Attention in Video Encoder.} We showed an example from TACoS dataset in Figure~\ref{fig:vidEncoderAtten}. For a given video snippet (\eg, the 30-\textit{th} frame\footnote{We slightly abuse "frame" and "patch" here, and we refer to them as a video snippet and a video cube, respectively. \label{frame_path}}), we selected two reference patches\footref{frame_path}, and visualized the attention weight heatmaps of the last self-attention layer on other four frames (\ie, the 10-\textit{th} frame to 50-\textit{th} frame). From Figure~\ref{fig:vidEncoderAtten}, we can observe that the self-attention in the video encoder can effectively focus more on the semantically corresponding areas (\eg, the person and chopping board) even across long temporal ranges, which is beneficial for encoding global context.
% \textit{s27-d29-cam-002}

\noindent\textbf{Cross-attention in Decoder.} An example from ANet dataset was presented in Figure~\ref{fig:DecoderAtten}. We took the stepwise fusion strategy for MCMA in the decoder. For the output, we selected the segment query slot which outputs the highest confidence scores and visualized its attention weights on two modal features. For the background video frames\footref{frame_path} (\#1), the decoder mainly focuses on figure outline. For ground-truth video frames\footref{frame_path} (\#2, \#3), it captures the most informative parts (\textit{e.g.,} moving hands and touching eye action), which plays an important role in location reasoning. As for the language attention weights, it focuses on essential words, \eg, the salient objects (lady, eye) and actions (put). %Note that we just visualize the results for the segment query slot with the highest confidence score.
%-------------------------------------------------------------------------------------%
%\input{tables_figs/figDecoderAtten} % Move before

%\input{tables_figs/tabSOTASplit}

%\input{imgs/umapVis} % Move to supply material
%\input{imgs/wordSelfAtten} % Move to supply material

%-------------------------------------------------------------------------------------%
%\input{tables/compare_sota1}
%\input{tables/compare_sota2}
%%\input{tables/compare_sota}
%\input{tables/tab3}
%-------------------------------------------------------------------------------------%
%\input{tables/end2endComp}
%-------------------------------------------------------------------------------------%

%-------------------------------------------------------------------------------------%
%\input{imgs/umapVis} % Move to supply material
%-------------------------------------------------------------------------------------%
%\input{imgs/wordSelfAtten} % Move to supply material
%-------------------------------------------------------------------------------------%
%%\input{imgs/vidEncoderAtten}
%-------------------------------------------------------------------------------------%
%%\input{imgs/DecoderAtten}

%-------------------------------------------------------------------------------------%
%\input{exps/4_5_vis} % Omit since it has been presented in 4_4_com_sota

\section{Conclusions and Future Works}
In this paper, we propose the first end-to-end multi-modal Transformer-family model GTR for video grounding. By carefully designing several novel components (\eg, cubic embedding layer, multi-head cross-modal attention module, and many-to-one matching loss), our GTR achieves new state-of-the-art performance on three challenging benchmarks. As a pioneering multi-modal Transformer model, we also conducted comprehensive explorations to summarize several empirical guidelines for model designs, which can help to open doors for the future research on multi-modal Transformers. Moving forward, we aim to extend GTR to more general settings, such as weakly-supervised video grounding or spatial-temporal video grounding.

\section{Acknowledgement}
This paper was partially supported by National Engineering Laboratory for Video Technology-Shenzhen Division, and Shenzhen Municipal Development and Reform Commission (Disciplinary Development Program for Data Science and Intelligent Computing). It is also supported by National Natural Science Foundation of China (NSFC 6217021843). Special acknowledgements are given to AOTO-PKUSZ Joint Research Center of Artificial Intelligence on Scene Cognition technology Innovation for its support. Mike Shou does not receive any funding for this work.

%\input{supplyMaterial}

\bibliography{refs.bib}
\bibliographystyle{acl_natbib}
\clearpage
%-----------------------------------------------------------------------
\section{Appendix}
\subsection{GTR Variant Settings}
We list typical GTR variant parameter settings in Table~\ref{tab:GTRvariant}.
%-----------------------------------------------------------------------
%\begin{wraptable}{r}{0.5\textwidth}
\begin{table}[h]
%\vspace{-12mm}
\small
\centering
\renewcommand\arraystretch{1.1}
\setlength{\tabcolsep}{3pt}{
\begin{tabular}{l|ccccc}
	\multicolumn{1}{l|}{Models}& $N_v$ & $N_s$ & $N_d$ & $d$  & Params(M)\\
	\shline
	GTR-B & 4 & 4  & 6 & 320 & 25.98  \\ 
	GTR-L & 6 & 6 & 8 & 320  & 40.56 \\
	GTR-H & 8 & 8 & 8 & 512  & 61.35 \\ 
	\hline
	GTR-B-Wide & 4 & 4  & 6 & 352 & 34.05  \\ 
	GTR-B-Deep & 4 & 4  & 8 & 320 & 33.89  \\ 
	\hline
	GTR-L-Wide & 6 & 6  & 8 & 352 & 47.18  \\ 
	GTR-L-Deep & 6 & 6  & 10 & 320 & 47.23  \\ 
\end{tabular}}
\vspace{-2mm}
\caption{\footnotesize{Typical GTR variant parameter settings.}}
\label{tab:GTRvariant}
\vspace{-5mm}
\end{table}
%\end{wraptable}
%-------------------------------------------------------------------------------------%
%-----------------------------------------------------------------------

%-----------------------------------------------------------------------
\subsection{Empirical Studies and Observations}\label{sec:ablation}
\subsubsection{Visual Token Acquisition}
Through ablation studies, we have found that temporal overlapping sampling plays a critical role in Cubic Embedding. Here we present more experimental results to confirm our findings. Specifically, we choose three kernel sizes (8, 8, 3), (16, 16, 3), and (8, 8, 6). In each case, we set five sets of stride shape corresponding to temporal overlapping, spatial overlapping, and non-overlapping cases, respectively. Experimental results in Table.~\ref{tab:cubicAbla} show that temporal overlapping brings about significant performance improvement under various kernel shape settings.

We also compare our Cubic Embedding layer with the framewise partition in Vision Transformer. Part of the results have been listed in the main paper and the full results are shown in Table.~\ref{tab:withViT}. It demonstrates the superiority of our cubic embedding layer.

%-----------------------------------------------------------------------
%-----------------------------------------------------------------------
%\begin{wraptable}{r}{0.5\textwidth}
\begin{table}[ht]
%\vspace{-12mm}
%\small
\centering
\renewcommand\arraystretch{1.1}
\resizebox{0.95\linewidth}{!}{
\begin{tabular}{c|c|ccc}
	Models &\multicolumn{1}{c|}{Stride}& $R_1^{0.5}$ & $R_1^{0.7}$  & GFLOPs\\
	\shline
	\rowcolor{gray!30} \multicolumn{5}{c}{kernel (8, 8, 3)} \\
	\hline
	\multirow{2}{*}{Temporal} & (8, 8, 1) & 49.73 & 28.51 & 41.05 \\
	~ & (8, 8, 2) & 49.67 & 28.45 & 20.55 \\
	\hline
	\multirow{2}{*}{Spatial} & (4, 4, 3) & 44.54 & 23.12 & 54.51 \\ 
	~ & (6, 6, 3) & 44.23 & 22.81 & 24.26\\ 
	\hline
	None & (8, 8, 3) & 43.73 & 22.45 & 14.69 \\ 
	\hline
	%  ----------------------------------------\\
	\rowcolor{gray!30} \multicolumn{5}{c}{kernel (16, 16, 3)} \\
	\hline
	\multirow{2}{*}{Temporal} & (16, 16, 1) & 43.23 & 24.13 & 10.83 \\
	~ & (16, 16, 2) & 43.14 & 23.93 & 5.44 \\
	\hline
	\multirow{2}{*}{Spatial} & (8, 8, 3) & 41.26 & 21.58 & 13.33 \\ 
	~ & (12, 12, 3) & 41.03 & 21.35 & 6.42 \\ 
	\hline
	None & (16, 16, 3) & 40.83 & 21.26 & 3.90 \\ 
	\hline
	%  ----------------------------------------
	\rowcolor{gray!30} \multicolumn{5}{c}{kernel (8, 8, 6)} \\
	\hline
	\multirow{2}{*}{Temporal} & (8, 8, 2) & 46.71 & 26.84 & 17.93 \\
	~ & (8, 8, 4) & 46.10 & 26.71 & 8.99 \\
	\hline
	\multirow{2}{*}{Spatial} & (4, 4, 6) & 42.96 & 21.93 & 22.21 \\ 
	~ & (6, 6, 6) & 42.95 & 21.90 & 9.90 \\ 
	\hline
	None & (8, 8, 6) & 42.84 & 21.61 & 6.01 \\ 
\end{tabular}
}
\vspace{-2mm}
\caption{\footnotesize{Input ablations for GTR-B on ActivityNet Caption dataset(\%).}}
\label{tab:cubicAbla}
\vspace{-5mm}
\end{table}
%\end{wraptable}
%-------------------------------------------------------------------------------------%

%-----------------------------------------------------------------------
%-----------------------------------------------------------------------
%-----------------------------------------------------------------------
%\begin{wraptable}{r}{0.6\textwidth}
\begin{table*}[th]
%	\vspace{-7mm}
	%\small
	\centering
	\renewcommand\arraystretch{1.1}
%	\scriptsize{
	\setlength{\tabcolsep}{3.5pt}{
		\resizebox{0.9\linewidth}{!}{
	\begin{tabular}{r|cccc|cccc|cccc}
	\multirow{2}{*}{Models} & \multicolumn{4}{c|}{ActivityNet Captions} & \multicolumn{4}{c|}{Charades-STA} & \multicolumn{4}{c}{TACoS} \\
	~  & $R_1^{0.5}$ & $R_1^{0.7}$  & $R_5^{0.5}$ & $R_5^{0.7}$ & $R_1^{0.5}$ & $R_1^{0.7}$  & $R_5^{0.5}$ & $R_5^{0.7}$  & $R_1^{0.3}$ & $R_1^{0.5}$  & $R_5^{0.3}$ & $R_5^{0.5}$ \\
	\shline
	Framewise & 46.10 & 26.27 & 76.51 & 62.04 & 60.05 & 36.94 & 89.40 & 60.25 & 35.98 & 26.34 & 57.09 & 45.72 \\
	Cubic & 49.67 & 28.45 & 79.83 & 64.34 & 62.45 & 39.23 & 91.40 & 61.76 & 39.34  & 28.34 & 60.85 & 46.67 \\
	%& 49.94 & 28.62 & 79.86 & 64.43 & 62.49 & 39.41 & 91.42 & 61.85 & 39.52 & 28.70 & 60.91  & 46.84 \\ 
	%Cubic & \textbf{50.43} & \textbf{28.74} & \textbf{80.14} & \textbf{64.60} & \textbf{62.50} & \textbf{39.50} & \textbf{91.58} & \textbf{61.93} & \textbf{39.60} & \textbf{29.04} & \textbf{61.04}  & \textbf{47.03} \\ 
	\end{tabular}}}%}
	\vspace{-2.5mm}
	\caption{\footnotesize{Framewise: ViT embedding manner; Cubic: ours.}}
	\label{tab:withViT}
	%\vspace{-4mm}
\end{table*}
%\end{wraptable} 
%-------------------------------------------------------------------------------------%

%-----------------------------------------------------------------------
%\section{Decoder Design Principles}\label{cubic}
%We focus on exploring whether the decoder should go deeper or wider? Based on the basic GTR-B model, we designed two GTR-B variants with nearly equivalent parameters: GTR-B-Wide and GTR-B-Deep. Specifically, GTR-B-Wide with the hidden dimension extended to 256 and GTR-B-Deep with layer number increased to 8. 

%-----------------------------------------------------------------------
\subsubsection{Decoder Design Principles}
%-----------------------------------------------------------------------

\noindent \textit{Columnar vs. Shrink vs. Expand}. We provide more experiments to determine how to design the attention head number in different layers.  We set up three different distributions of attention heads: i) the same number of heads in all layers (columnar); ii) gradually decrease the number of heads (shrink); iii) gradually increase the number of heads (expand). The results are listed in Table.~\ref{tab:expand}. It is consistent with our conclusion that \textbf{progressively expanding the attention head leads to better performance}. %Interestingly, it is consistent with the ResNet designing rule (\ie, enlarging feature channels with the growth of network depth).

%-----------------------------------------------------------------------
%-----------------------------------------------------------------------
%\begin{wraptable}{r}{0.6\textwidth}
\begin{table*}[th]
%	\vspace{-7mm}
	%\small
	\centering
	\renewcommand\arraystretch{1.1}
%	\scriptsize{
	\setlength{\tabcolsep}{3.5pt}{
		\resizebox{\linewidth}{!}{
	\begin{tabular}{r|c|cccc|cccc|cccc}
	\multirow{2}{*}{Models} & \multirow{2}{*}{Heads} & \multicolumn{4}{c|}{ActivityNet Captions} & \multicolumn{4}{c|}{Charades-STA} & \multicolumn{4}{c}{TACoS} \\
	~ & ~ & $R_1^{0.5}$ & $R_1^{0.7}$  & $R_5^{0.5}$ & $R_5^{0.7}$ & $R_1^{0.5}$ & $R_1^{0.7}$  & $R_5^{0.5}$ & $R_5^{0.7}$  & $R_1^{0.3}$ & $R_1^{0.5}$  & $R_5^{0.3}$ & $R_5^{0.5}$ \\
	\shline
	\multirow{2}{*}{GTR-B-Columnar} & [4, 4, 4, 4, 4, 4] & 49.42 & 28.34 & 79.75 & 64.20  & 62.34 & 39.02 & 91.25 & 61.41 & 39.01 & 27.93 & 60.42  & 46.39 \\ 
	~ & [8, 8, 8, 8, 8, 8] & 49.43 & 28.34 & 79.76 & 64.23  & 62.36 & 39.03 & 91.26 & 61.41 & 39.03 & 27.93 & 60.42  & 46.39 \\ 
	\hline
	\multirow{2}{*}{GTR-B-Shrink}  & [8, 8, 4, 4, 2, 1] & 49.01 & 27.95 & 79.26 & 64.00  & 62.15 & 38.81 & 90.93 & 61.45 & 38.95 & 27.73 & 59.93  & 45.92 \\ 
	~ & [16, 16, 8, 8, 4, 1]  & 49.03 & 27.96 & 79.26 & 64.02  & 62.16 & 38.81 & 90.93 & 61.45 & 38.95 & 27.74 & 59.94  & 45.92 \\ 
	\hline
	\multirow{2}{*}{GTR-B-Expand} & [1, 2, 4, 4, 8, 8] & 49.67 & 28.45  & 79.83 & 64.34  & 62.45 & 39.23 & 91.40 & 61.76  & 39.34  & 28.34 & 60.85 & 46.67 \\ 
	~ & [1, 4, 8, 8, 16, 16] & 49.64 & 28.42  & 79.81 & 64.32  & 62.45 & 39.22 & 91.41 & 61.77  & 39.33  & 28.33 & 60.85 & 46.67 \\ 	
	\end{tabular}}}%}
	\vspace{-2.5mm}
	\caption{\footnotesize{Columnar vs. Shrink vs. Expand. Progressively expanding the attention heads leads to better performance.}}
	\label{tab:expand}
	%\vspace{-4mm}
\end{table*}
%\end{wraptable} 
%-------------------------------------------------------------------------------------%

%-----------------------------------------------------------------------
%-----------------------------------------------------------------------
%\begin{wraptable}{r}{0.5\textwidth}
\begin{table*}
%\vspace{-2mm}
\small
\centering
\renewcommand\arraystretch{1.1}
\setlength{\tabcolsep}{3pt}{
\begin{tabular}{x{50}|x{40}|x{35}x{35}x{35}x{35}}
Sample Rate & Crop size & $R_1^{0.5}$  & $R_1^{0.7}$  & $R_5^{0.5}$  & $R_5^{0.7}$ \\
\shline
\multirow{4}{*}{1/8} & 112 & 49.67 & 28.45 & 79.83 & 64.34 \\
~ & 140 & 49.72 & 28.52 & 79.95 & 64.40 \\
~ & 168 & 49.73 & 28.52 & 79.97 & 64.40 \\
~ & 224 & 49.73 & 28.52 & 79.97 & 64.40 \\
\hline
1/8 & \multirow{4}{*}{112} & 49.67 & 28.45 & 79.83 & 64.34 \\
1/6 & ~ & 51.54 & 29.43 & 80.11 & 64.76 \\
1/4 & ~ & 52.34 & 29.57 & 80.14 & 64.79 \\
1/2 & ~ & 52.35 & 29.57 & 80.14 & 64.79 \\
\end{tabular}
}
\vspace{-3mm}
\caption{\footnotesize{Performance (\%) on ANet on different frame crop size and sample rate.}}
\label{tab:vsCropSample}
%\vspace{-3mm}
\end{table*}
%\end{wraptable}
%-------------------------------------------------------------------------------------%

\noindent \textit{Deep vs. Wide}. We have developed two variants (GTR-B-Wide, GTR-B-Deep) and reported the results on ActivityNet Caption and TACoS datasets, which demonstrates that going deeper is more effective than going wider. Here we provide two additional variants (GTR-L-Wide, GTR-L-Deep) and report results on all three datasets in Table~\ref{tab:deepwide}. Similarly, the deep model also outperforms the wide model, which further confirms our conclusion.
%-----------------------------------------------------------------------
%-----------------------------------------------------------------------
%\begin{wraptable}{r}{0.6\textwidth}
\begin{table*}[th]
%	\vspace{-7mm}
	%\small
	\centering
	\renewcommand\arraystretch{1.1}
%	\scriptsize{
	\setlength{\tabcolsep}{3.5pt}{
		\resizebox{0.9\linewidth}{!}{
	\begin{tabular}{l|c|cccc|cccc|cccc}
	\multirow{2}{*}{Models} & \multirow{2}{*}{Param(M)} & \multicolumn{4}{c|}{ActivityNet Captions} & \multicolumn{4}{c|}{Charades-STA} & \multicolumn{4}{c}{TACoS} \\
	~ & ~ & $R_1^{0.5}$ & $R_1^{0.7}$  & $R_5^{0.5}$ & $R_5^{0.7}$ & $R_1^{0.5}$ & $R_1^{0.7}$  & $R_5^{0.5}$ & $R_5^{0.7}$  & $R_1^{0.3}$ & $R_1^{0.5}$  & $R_5^{0.3}$ & $R_5^{0.5}$ \\
	\shline
	GTR-B-Wide & 34.05 & 49.94 & 28.62 & 79.86 & 64.43 & 62.49 & 39.41 & 91.42 & 61.85 & 39.52 & 28.70 & 60.91  & 46.84 \\ 
	GTR-B-Deep & 33.89 & \textbf{50.15} & \textbf{28.74} & \textbf{80.14} & \textbf{64.60} & \textbf{62.50} & \textbf{39.50} & \textbf{91.58} & \textbf{61.93} & \textbf{39.60} & \textbf{29.04} & \textbf{61.04}  & \textbf{47.03} \\ 
	\hline
	GTR-L-Wide & 47.18 & 50.48 & 28.96 & 80.28 & 65.00 & 62.51 & 39.58 & 91.62 & 61.94 & 39.98 & 29.35 & 61.45  & 47.30 \\ 
	GTR-L-Deep & 47.23 & \textbf{50.52} & \textbf{29.05} & \textbf{80.37} & \textbf{65.03} & \textbf{62.54} & \textbf{39.66} & \textbf{91.62} & \textbf{62.00} & \textbf{40.12} & \textbf{29.51} & \textbf{61.75}  & \textbf{47.52} \\ 
	\end{tabular}}}%}
	\vspace{-2.5mm}
	\caption{\footnotesize Performance (\%) comparisons between the wide variants and deep variants.}
	\label{tab:deepwide}
	%\vspace{-4mm}
\end{table*}
%\end{wraptable} 
%-------------------------------------------------------------------------------------%

%-----------------------------------------------------------------------

%-----------------------------------------------------------------------
%\begin{wraptable}{r}{0.6\textwidth}
\begin{table*}[ht]
%	\vspace{-7mm}
	%\small
	\centering
	\renewcommand\arraystretch{1.1}
%	\scriptsize{
	\setlength{\tabcolsep}{3.5pt}{
		\resizebox{\linewidth}{!}{
	\begin{tabular}{r|cccc|cccc|cccc|cc}
	\multirow{2}{*}{Models} & \multicolumn{4}{c|}{ActivityNet Captions} & \multicolumn{4}{c|}{Charades-STA} & \multicolumn{4}{c|}{TACoS} & \multirow{2}{*}{Param(M)} & \multirow{2}{*}{\small{GFLOPs}}\\
	~ & $R_1^{0.5}$ & $R_1^{0.7}$  & $R_5^{0.5}$ & $R_5^{0.7}$ & $R_1^{0.5}$ & $R_1^{0.7}$  & $R_5^{0.5}$ & $R_5^{0.7}$  & $R_1^{0.3}$ & $R_1^{0.5}$  & $R_5^{0.3}$ & $R_5^{0.5}$ & ~ & ~ \\
	\shline
	Early Fusion & 39.35 & 19.64 & 70.39 & 56.34  & 54.13 & 32.72 & 84.69 & 54.53 & 28.25 & 19.56 & 50.46  & 36.06 & 12.86 & 11.31\\ 
	Conditional Fusion & 35.25 & 14.57 & 66.35 & 51.34  & 50.62 & 25.65 & 79.34 & 49.03 & 24.62 & 14.37 & 47.34  & 34.74 & 12.42 & 10.60\\ 
	\hline
	Joint Fusion & 42.51 & 21.53 & 71.24 & 58.04  & 57.66 & 33.68 & 87.52 & 57.34 & 33.69 & 23.72 & 54.27  & 40.14 & 16.62 & 13.91\\ 
	Divided Fusion   & 46.91 & 26.21 & 76.85 & 63.13  & 60.93 & 38.22 & 89.84 & 59.95 & 37.82 & 26.31 & 58.13  & 44.80 & 25.16 & 21.02\\ 
	Hybrid Fusion & \textcolor{blue}{\textbf{46.82}} & \textcolor{blue}{\textbf{25.45}}  &  \textcolor{blue}{\textbf{76.20}} & \textcolor{blue}{\textbf{62.41}} & \textcolor{blue}{\textbf{60.21}} & \textcolor{blue}{\textbf{37.25}}& \textcolor{blue}{\textbf{89.63}} & \textcolor{blue}{\textbf{59.51}}& \textcolor{blue}{\textbf{37.01}}& \textcolor{blue}{\textbf{25.77}} &\textcolor{blue}{\textbf{57.63}}& \textcolor{blue}{\textbf{44.32}} & \textcolor{blue}{\textbf{20.94}} & \textcolor{blue}{\textbf{17.25}} \\
	\footnotesize{Hybrid Fusion(value split)} & 46.69 & 25.32 & 75.83 & 61.93  & 59.94 & 36.93 & 89.52 & 59.32 & 36.85 & 25.49 & 57.42  & 44.05 & 20.82 & 16.75\\ 
	Stepwise Fusion & \textcolor{red}{\textbf{49.67}} &  \textcolor{red}{\textbf{28.45}}  & \textcolor{red}{\textbf{79.83}} & \textcolor{red}{\textbf{64.34}}  & \textcolor{red}{\textbf{62.45}} & \textcolor{red}{\textbf{39.23}}& \textcolor{red}{\textbf{91.40}} & \textcolor{red}{\textbf{61.76}}  & \textcolor{red}{\textbf{39.34}}  & \textcolor{red}{\textbf{28.34}} & \textcolor{red}{\textbf{60.85}} &\textcolor{red}{\textbf{46.67}} & \textcolor{red}{\textbf{25.98}} & \textcolor{red}{\textbf{20.55}}\\
	\footnotesize{Stepwise Fusion(L-V)} & 49.62 & 28.37 & 79.76 & 64.25 & 62.37 & 38.93 & 91.17 & 61.53 & 39.19 & 28.29 & 60.62  & 46.55 & 25.98 & 20.55\\ 
	\end{tabular}}}%}
	\vspace{-2.5mm}
	\caption{\footnotesize{Multi-modal fusion comparisons.}}
	\label{tab:fuseVarient}
	%\vspace{-4mm}
\end{table*}
%\end{wraptable} 
%-------------------------------------------------------------------------------------%

%-----------------------------------------------------------------------

%-----------------------------------------------------------------------
\subsubsection{Multi-modal Fusion Mechanisms}
\begin{figure*}[t]
	\centering
	%	\begin{subfigure}[b]{\textwidth}
	%		\centering
	\includegraphics[width=\textwidth]{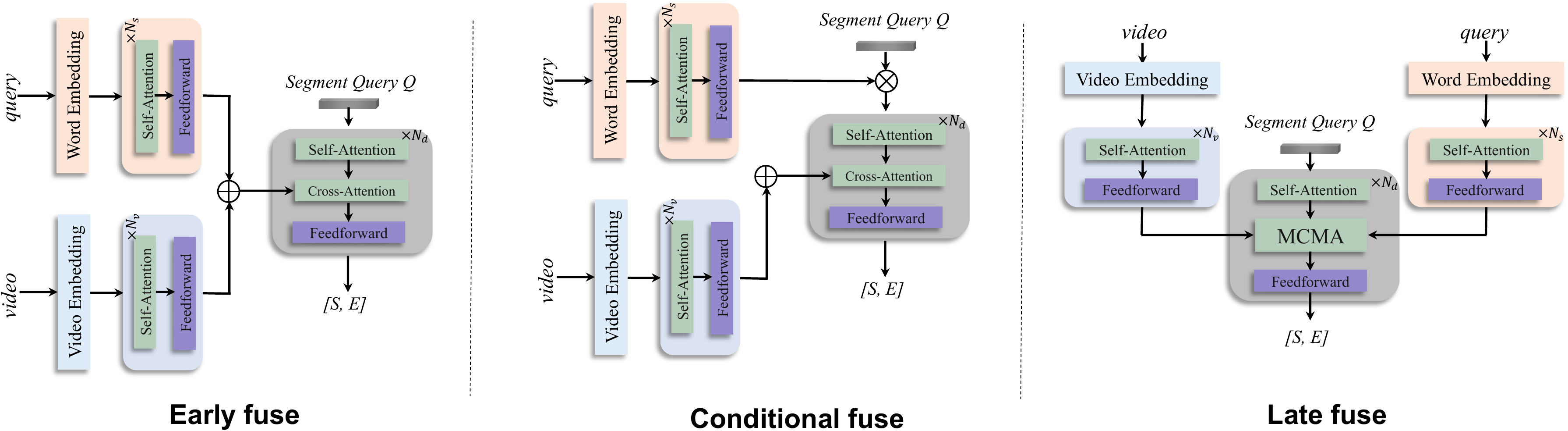}
	%		\caption{$m=0$}
	%	\end{subfigure}
	%	\begin{subfigure}[b]{0.9\textwidth}
	%\centering
	%\includegraphics[width=\textwidth]{figs/fuse.pdf}
	%\caption{$m=1$}
	%	\end{subfigure}
	\caption{\footnotesize Three methodologies of multi-modal fusion.}
	\label{fig:threeFuse}
\end{figure*}

%-----------------------------------------------------------------------
In general, multi-modal fusion methods can be divided into three categories (cf. Figure.~\ref{fig:threeFuse}): 1) \emph{Early fusion:} The multi-modal features are fused before being fed into the decoder. 2) \emph{Conditional fusion}: The language feature acts as conditional signals of segment queries of the decoder. We concatenate it with the learnable segment queries features. 3) \emph{Late fusion:} Multi-modal fusion is conducted with the decoder via the proposed Multi-head Cross-modal Attention (MCMA) Module.
%-----------------------------------------------------------------------
%-----------------------------------------------------------------------

\begin{figure*}[t]
	\centering
	\includegraphics[width=\textwidth]{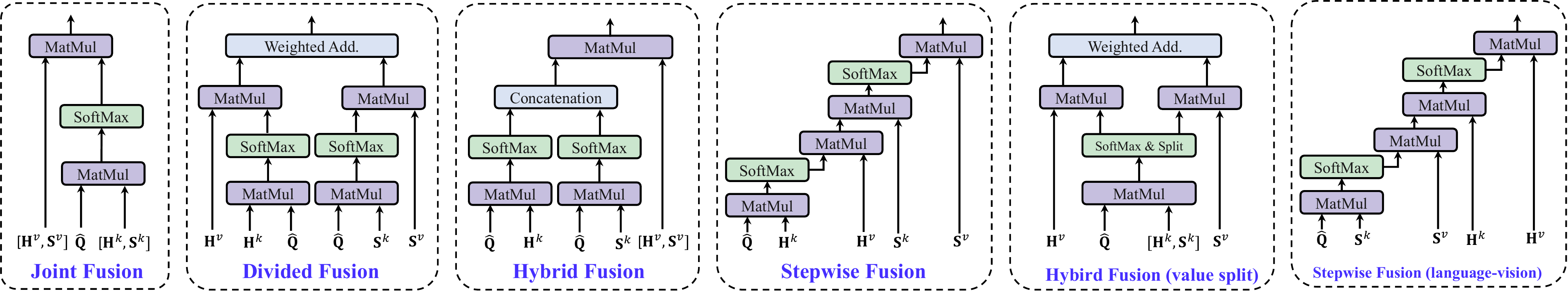}
	\vspace{-3mm}
	\caption{\footnotesize{Modal variants of  Multi-head Cross-Modal Attention module (MCMA). The first fourth variants are already presented in the main paper and the rightmost two are two variants for Hybrid Fusion and Stepwise Fusion respectively.}}
	\label{fig:fuseFull}
\end{figure*}

For the specific instantiations of MCMA, except for the four fusion strategies mentioned in the main paper, we additionally present two additional variants in Figure.~\ref{fig:fuseFull}. \textbf{1)} \textit{Hybrid Fusion (value split)} concatenates the key features and computes the query and value multiplication separately. Specifically, $\mathbf{\hat{Q}} \in \mathbb{R}^{N \times d}$, $\mathbf{H}^k  \in \mathbb{R}^{F \times d}$, $\mathbf{S}^k  \in \mathbb{R}^{S \times d}$ generate the attention matrix with shape $(F+S) \times d$, which is divided into two parts and each is with shape $F \times d$ and $S \times d$, respectively. These two divided attention weights are used to computed with $\mathbf{H}^v$ and $\mathbf{S}^v$ separately to generate the final results. \textbf{2)} \textit{Stepwise Fusion (language-vision)} fuses the language feature firstly and then the video features. 

The experimental results are presented in Table.~\ref{tab:fuseVarient} and we have the following findings. 1) Stepwise fusion has the best performance on all three datasets by using the largest number of parameters. 2) The performance of hybrid fusion (value split) is almost the same as hybrid fusion. 3) Also, stepwise fusion(L-V) shares the similar performance with stepwise fusion, which demonstrates that the order of fusion of visual or language information is not sensitive.

To investigate the influence of frame crop size and sample rate, we conduct more experiments on stepwise fusion models. The results in Table.~\ref{tab:vsCropSample} show that 1) the performance is not sensitive o the frame crop size; 2) the performance is positively correlated with the sample rate but gradually reach convergence.
%-----------------------------------------------------------------------
%\input{Appendix/tabVSCropSample}
%-----------------------------------------------------------------------
%-----------------------------------------------------------------------
%\textcolor{red}{Specify the differences between early fuse and joint fuse.}

\subsection{Performance of GTR}\label{sec:SOTA}
The performance of GTR under more IoUs is available in Table.~\ref{tab:fullPerform}.
%-----------------------------------------------------------------------
%-----------------------------------------------------------------------
%\begin{wraptable}{r}{0.6\textwidth}
\begin{table*}[ht]
	%\vspace{-7mm}
	\small
	\centering
	\renewcommand\arraystretch{1.1}
%	\scriptsize{
	\setlength{\tabcolsep}{3.5pt}{
		\resizebox{\linewidth}{!}{
	\begin{tabular}{l|cccccc|cccc|cccccc}
	\multirow{2}{*}{Models} & \multicolumn{6}{c|}{ActivityNet Captions} & \multicolumn{4}{c|}{Charades-STA} & \multicolumn{6}{c}{TACoS}\\
	~ & $R_1^{0.3}$ & $R_1^{0.5}$ & $R_1^{0.7}$  & $R_1^{0.3}$ & $R_5^{0.5}$ & $R_5^{0.7}$ & $R_1^{0.5}$ & $R_1^{0.7}$  & $R_5^{0.5}$ & $R_5^{0.7}$ & $R_1^{0.1}$ & $R_1^{0.3}$ & $R_1^{0.5}$ & $R_1^{0.1}$ & $R_5^{0.3}$ & $R_5^{0.5}$ \\
	\shline
	GTR-B (Ours)  & {\color{blue}\textbf{66.15}}  & {\color{blue}\textbf{49.67}} & {\color{blue}\textbf{28.45}} & {\color{blue}\textbf{87.94}} & {\color{blue}\textbf{79.83}} & {\color{blue}\textbf{64.34}} & \color{blue}\textbf{62.45} & \color{blue}\textbf{39.23}&  \color{blue}\textbf{91.40} & \color{blue}\textbf{61.76} & {\color{blue}\textbf{48.85}} & \color{blue}\textbf{39.34} & \color{blue}\textbf{28.34} & {\color{blue}\textbf{72.59}} & \color{blue}\textbf{60.85}&  \color{blue}\textbf{46.67}  \\
	GTR-L (Ours) & {\textbf{66.72}} & {\textbf{50.43}} & {\textbf{28.91}} & {\textbf{88.21}} & {\textbf{80.22}} & {\textbf{64.95}}  & \textbf{62.51} & \textbf{39.56}&  \textbf{91.62} & \textbf{61.97} & {\textbf{49.25}} & \textbf{39.93} & \textbf{29.21} & {\textbf{73.14}} & \textbf{61.22}&  \textbf{47.10}  \\
	GTR-H (Ours) & {\color{red}\textbf{66.80}} & {\color{red}\textbf{50.57}} & {\color{red}\textbf{29.11}} & {\color{red}\textbf{88.34}} & {\color{red}\textbf{80.43}} & {\color{red}\textbf{65.14}} & \color{red}\textbf{62.58} & \color{red}\textbf{39.68}&  \color{red}\textbf{91.62} & \color{red}\textbf{62.03} & {\color{red}\textbf{49.41}} & \color{red}\textbf{40.39} & \color{red}\textbf{30.22} & {\color{red}\textbf{73.24}} & \color{red}\textbf{61.94} & \color{red}\textbf{47.73}   \\
	\end{tabular}}}%}
	\vspace{-2.5mm}
	%\caption{\footnotesize Performance comparisons on three benchmarks(\%). {\color{blue}\textbf{GTR-B}} is more efficient while {\color{red}\textbf{GTR-H}} achieves the highest recall.}
	\caption{\footnotesize Performance comparisons on three benchmarks(\%).}
	\label{tab:fullPerform}
	\vspace{-2mm}
\end{table*}
%\end{wraptable} 
%-------------------------------------------------------------------------------------%
%-----------------------------------------------------------------------

%\appendix

%\section{Example Appendix}
%\label{sec:appendix}

%This is an appendix.

\end{document}